\documentclass[ijoc,nonblindrev]{informs3} 
\OneAndAHalfSpacedXI 

\usepackage{endnotes}
\let\footnote=\endnote
\let\enotesize=\normalsize
\def\notesname{Endnotes}%
\def\makeenmark{$^{\theenmark}$}
\def\enoteformat{\rightskip0pt\leftskip0pt\parindent=1.75em
  \leavevmode\llap{\theenmark.\enskip}}

\usepackage[dvipsnames]{xcolor}
\usepackage{mwe}
\usepackage{tikz}
\usepackage[abs]{overpic}
\usetikzlibrary{arrows.meta,decorations.pathreplacing,angles,quotes}
\usepackage{longtable}
\usepackage{enumerate}
\usepackage{mathrsfs,amsfonts}
\usepackage{hyperref}
\usepackage{lscape}
\usepackage{standalone}
\usepackage{caption}
\usepackage{subcaption}
\usepackage[absolute,overlay]{textpos}
\usepackage{dsfont}
\usepackage[margin=1in]{geometry}
\usepackage{booktabs}
\usepackage{comment}
\usepackage{transparent}
\usepackage[normalem]{ulem}
\usepackage{mathtools}
\usepackage{multirow}
\usepackage{multicol}
\usepackage{float}
\usepackage{arydshln}
\usepackage{pgfplots}
\pgfplotsset{compat=newest,ticks=none}
\usepgfplotslibrary{patchplots}
\usepackage{makecell}
\usepackage{enumitem}
\usepackage{multicol}
\usepackage{romannum}
\usepackage{pgfplots}
\usepgfplotslibrary{fillbetween}
\pgfplotsset{compat=1.18}
\usepackage[noend]{algpseudocode}
\usepackage[absolute,overlay]{textpos}
\usepackage{algorithm}
\usepackage{algpseudocode}
\usepackage[left,pagewise,mathlines]{lineno}\nolinenumbers

\newlist{deflist}{description}{2}
\setlist[deflist]{labelwidth=2cm,leftmargin=!,font=\normalfont}

\newcommand{\soct}{\operatorname{S-OCT}}
\newcommand{\octh}{\operatorname{OCT-H}}
\newcommand{\cart}{\operatorname{CART}}

\newcommand{\mcutone}{\operatorname{CUT_{w}-H}}
\newcommand{\mcuttwo}{\operatorname{CUT-H}}

\newcommand{\child}{\operatorname{CHILD}}

\newcommand{\milo}{\operatorname{MILO}}

\newcommand{\dist}{\operatorname{dist}}

\usepackage{natbib}
 \bibpunct[, ]{(}{)}{,}{a}{}{,}%
 \def\bibfont{\small}%
\TheoremsNumberedThrough     
\ECRepeatTheorems

\EquationsNumberedThrough    

\begin{document}


\RUNAUTHOR{Alston, Validi, \& Hicks}

\RUNTITLE{MILO formulations for multivariate classification trees}

\TITLE{Optimal Mixed Integer Linear Optimization Trained Multivariate Classification Trees}
\ARTICLEAUTHORS{
\AUTHOR{Brandon Alston}
\AFF{Computational Applied Mathematics and Operations Research, Rice University, Houston, TX 77005,
\EMAIL{bca3@rice.edu}}
\AUTHOR{Illya V. Hicks}
\AFF{Computational Applied Mathematics and Operations Research, Rice University, Houston, TX 77005,
\EMAIL{ivhicks@rice.edu}}
}

\ABSTRACT{Multivariate decision trees are powerful machine learning tools for classification and regression that attract many researchers and industry professionals. An optimal binary tree has two types of vertices, (i) branching vertices which have exactly two children and where datapoints are assessed on a set of discrete features and (ii) leaf vertices at which datapoints are given a prediction, and can be obtained by solving a biobjective optimization problem that seeks to (i) maximize the number of correctly classified datapoints and (ii) minimize the number of branching vertices. Branching vertices are linear combinations of training features and therefore can be thought of as hyperplanes. In this paper, we propose two cut-based mixed integer linear optimization (MILO) formulations for designing optimal binary classification trees (leaf vertices assign discrete classes). Our models leverage on-the-fly identification of minimal infeasible subsystems (MISs) from which we derive cutting planes that hold the form of packing constraints. We show theoretical improvements on the strongest flow-based MILO formulation currently in the literature and conduct experiments on publicly available datasets to show our models' ability to scale, strength against traditional branch and bound approaches, and robustness in out-of-sample test performance. Our code and data are available on GitHub.}



\maketitle

\section{Introduction}
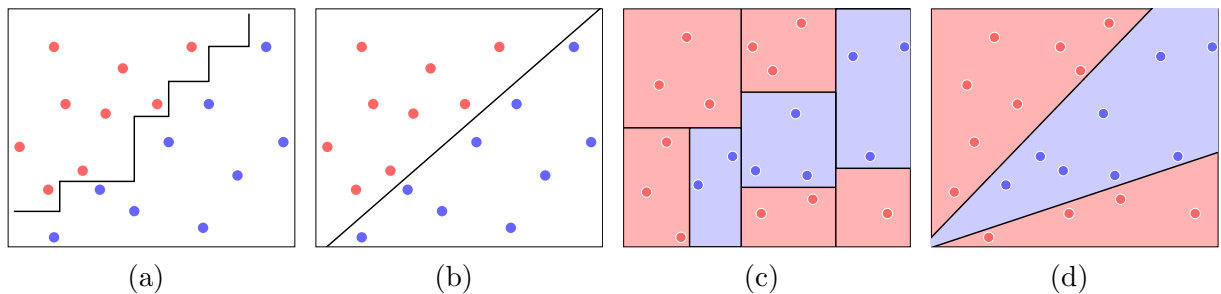
\begin{figure}[H]
\begin{subfigure}{.24\linewidth}
    \centering
    \begin{tikzpicture}
    \begin{axis}[ticks=none,
        xmin=0,xmax=5,
        ymin=0,ymax=5,
        enlargelimits=false,
        xlabel style={at={(ticklabel* cs:1)},anchor=north west},
        ylabel style={at={(ticklabel* cs:1)},anchor=south west}]
    \draw [line width=0.35mm] (0.1,.745) -- (.9,.745) -- (.9, 1.375) -- (2.2,1.375) -- (2.2, 2.11) -- (2.2, 2.74) -- (2.8, 2.74) -- (2.8, 3.475) -- (3.5, 3.475) -- (3.5, 4.21) -- (4.2, 4.21) -- (4.2,4.9);
    \node[circle, fill=blue!60, inner sep=2.5pt] at (axis cs:.8,.2) {};
    \node[circle, fill=blue!60, inner sep=2.5pt] at (axis cs:1.6,1.2) {};
    \node[circle, fill=blue!60, inner sep=2.5pt] at (axis cs:4.5,4.2) {};
    \node[circle, fill=blue!60, inner sep=2.5pt] at (axis cs:3.5,3) {};
    \node[circle, fill=blue!60, inner sep=2.5pt] at (axis cs:4.8,2.2) {};
    \node[circle, fill=blue!60, inner sep=2.5pt] at (axis cs:4,1.5) {};
    \node[circle, fill=blue!60, inner sep=2.5pt] at (axis cs:2.8,2.2) {};
    \node[circle, fill=blue!60, inner sep=2.5pt] at (axis cs:2.2,.75) {};
    \node[circle, fill=blue!60, inner sep=2.5pt] at (axis cs:3.4,.4) {};
    \node[circle, fill=red!60, inner sep=2.5pt] at (axis cs:.7,1.2) {};
    \node[circle, fill=red!60, inner sep=2.5pt] at (axis cs:1.3,1.6) {};
    \node[circle, fill=red!60, inner sep=2.5pt] at (axis cs:2,3.75) {};
    \node[circle, fill=red!60, inner sep=2.5pt] at (axis cs:3.2,4.2) {};
    \node[circle, fill=red!60, inner sep=2.5pt] at (axis cs:1.7,2.8) {};
    \node[circle, fill=red!60, inner sep=2.5pt] at (axis cs:.8,4.2) {};
    \node[circle, fill=red!60, inner sep=2.5pt] at (axis cs:.2,2.1) {};
    \node[circle, fill=red!60, inner sep=2.5pt] at (axis cs:2.6,3) {};
    \node[circle, fill=red!60, inner sep=2.5pt] at (axis cs:1,3) {};
    \end{axis}
\end{tikzpicture}
    \caption{}
\end{subfigure}
\begin{subfigure}{.24\linewidth}
    \centering
    \begin{tikzpicture}
    \begin{axis}[ticks=none,
        xmin=0,xmax=5,
        ymin=0,ymax=5,
        enlargelimits=false,
        xlabel style={at={(ticklabel* cs:1)},anchor=north west},
        ylabel style={at={(ticklabel* cs:1)},anchor=south west}]

    \draw[domain=0:5, smooth, variable=\x, line width=.35mm] plot ({\x}, {1.05*\x-0.2});

    \node[circle, fill=blue!60, inner sep=2.5pt] at (axis cs:.8,.2) {};
    \node[circle, fill=blue!60, inner sep=2.5pt] at (axis cs:1.6,1.2) {};
    \node[circle, fill=blue!60, inner sep=2.5pt] at (axis cs:4.5,4.2) {};
    \node[circle, fill=blue!60, inner sep=2.5pt] at (axis cs:3.5,3) {};
    \node[circle, fill=blue!60, inner sep=2.5pt] at (axis cs:4.8,2.2) {};
    \node[circle, fill=blue!60, inner sep=2.5pt] at (axis cs:4,1.5) {};
    \node[circle, fill=blue!60, inner sep=2.5pt] at (axis cs:2.8,2.2) {};
    \node[circle, fill=blue!60, inner sep=2.5pt] at (axis cs:2.2,.75) {};
    \node[circle, fill=blue!60, inner sep=2.5pt] at (axis cs:3.4,.4) {};

    \node[circle, fill=red!60, inner sep=2.5pt] at (axis cs:.7,1.2) {};
    \node[circle, fill=red!60, inner sep=2.5pt] at (axis cs:1.3,1.6) {};
        \node[circle, fill=red!60, inner sep=2.5pt] at (axis cs:2,3.75) {};
    \node[circle, fill=red!60, inner sep=2.5pt] at (axis cs:3.2,4.2) {};
    \node[circle, fill=red!60, inner sep=2.5pt] at (axis cs:1.7,2.8) {};
    \node[circle, fill=red!60, inner sep=2.5pt] at (axis cs:.8,4.2) {};
    \node[circle, fill=red!60, inner sep=2.5pt] at (axis cs:.2,2.1) {};
    \node[circle, fill=red!60, inner sep=2.5pt] at (axis cs:2.6,3) {};
    \node[circle, fill=red!60, inner sep=2.5pt] at (axis cs:1,3) {};
    \end{axis}
\end{tikzpicture}
    \caption{}
\end{subfigure}
\begin{subfigure}{.24\linewidth}
    \centering
    \begin{tikzpicture}
    \begin{axis}[ticks=none,
        xmin=0,xmax=5,
        ymin=0,ymax=5,
        enlargelimits=false,
        xlabel style={at={(ticklabel* cs:1)},anchor=north west},
        ylabel style={at={(ticklabel* cs:1)},anchor=south west},
        axis background/.style={fill=red!30}]
    \draw [fill=blue!20, line width=.35mm] (5,5) -- (3.7,5) -- (3.7,1.65) -- (5,1.65);
    \draw [fill=blue!20, line width=.35mm] (3.7,1.25) -- (2.05,1.25) -- (2.05,3.25) -- (3.7,3.25);
    \draw [fill=blue!20, line width=.35mm] (2.05,0) -- (2.05,2.5) -- (1.15,2.5) -- (1.15,0);
    \draw [line width=.35mm] (3.7,3.25) -- (3.7,1.25);
    \draw [line width=.35mm] (3.7,1.25) -- (3.7,0);
    \draw [line width=.35mm] (2.05,3.25) -- (2.05,5);
    \draw [line width=.35mm] (1.15,2.5) -- (0,2.5);
    \draw [line width=.35mm] (1.15,0) -- (2.05,0);

    \node[circle, draw=white, fill=blue!60, inner sep=2.5pt] at (axis cs:1.3,1.3) {};
    \node[circle, draw=white, fill=blue!60, inner sep=2.5pt] at (axis cs:3,2.8) {};
    \node[circle, draw=white, fill=blue!60, inner sep=2.5pt] at (axis cs:4,4) {};
    \node[circle, draw=white, fill=blue!60, inner sep=2.5pt] at (axis cs:4.9,4.2) {};
    \node[circle, draw=white, fill=blue!60, inner sep=2.5pt] at (axis cs:3.2,1.5) {};
    \node[circle, draw=white, fill=blue!60, inner sep=2.5pt] at (axis cs:2.3,1.6) {};
    \node[circle, draw=white, fill=blue!60, inner sep=2.5pt] at (axis cs:4.3,1.9) {};
    \node[circle, draw=white, fill=blue!60, inner sep=2.5pt] at (axis cs:1.9,1.9) {};

    \node[circle, draw=white, fill=red!60, inner sep=2.5pt] at (axis cs:1, 0.2) {};
    \node[circle, draw=white, fill=red!60, inner sep=2.5pt] at (axis cs:0.4, 1.15) {};
    \node[circle, draw=white, fill=red!60, inner sep=2.5pt] at (axis cs:3.3, 1) {};
    \node[circle, draw=white, fill=red!60, inner sep=2.5pt] at (axis cs:4.6, .7) {};
    \node[circle, draw=white, fill=red!60, inner sep=2.5pt] at (axis cs:2.6, 3.7) {};
    \node[circle, draw=white, fill=red!60, inner sep=2.5pt] at (axis cs:3.1, 4.7) {};
    \node[circle, draw=white, fill=red!60, inner sep=2.5pt] at (axis cs:1.1, 4.4) {};
    \node[circle, draw=white, fill=red!60, inner sep=2.5pt] at (axis cs:2.25, 4.2) {};
    \node[circle, draw=white, fill=red!60, inner sep=2.5pt] at (axis cs:2.25, 4.2) {};
    \node[circle, draw=white, fill=red!60, inner sep=2.5pt] at (axis cs:0.75, 2.2) {};
    \node[circle, draw=white, fill=red!60, inner sep=2.5pt] at (axis cs:1.5, 3) {};
    \node[circle, draw=white, fill=red!60, inner sep=2.5pt] at (axis cs:.62, 3.4) {};
    \node[circle, draw=white, fill=red!60, inner sep=2.5pt] at (axis cs:2.4, .7) {};

    \end{axis}
\end{tikzpicture}
    \caption{}
\end{subfigure}
\begin{subfigure}{.24\linewidth}
    \centering
    \begin{tikzpicture}
    \begin{axis}[ticks=none,
        xmin=0,xmax=5,
        ymin=0,ymax=5,
        enlargelimits=false,
        xlabel style={at={(ticklabel* cs:1)},anchor=north west},
        ylabel style={at={(ticklabel* cs:1)},anchor=south west},
        axis background/.style={fill=red!30}]
    \addplot[smooth, line width=.35mm, name path=A] {0.4*x-.015}; 
    \addplot[smooth, line width=.35mm, name path=B] {1.25*x+.2}; 
    \addplot[blue!20] fill between[of=A and B]; 

    \node[circle, draw=white, fill=blue!60, inner sep=2.5pt] at (axis cs:1.3,1.3) {};
    \node[circle, draw=white, fill=blue!60, inner sep=2.5pt] at (axis cs:3,2.8) {};
    \node[circle, draw=white, fill=blue!60, inner sep=2.5pt] at (axis cs:4,4) {};
    \node[circle, draw=white, fill=blue!60, inner sep=2.5pt] at (axis cs:4.9,4.2) {};
    \node[circle, draw=white, fill=blue!60, inner sep=2.5pt] at (axis cs:3.2,1.5) {};
    \node[circle, draw=white, fill=blue!60, inner sep=2.5pt] at (axis cs:2.3,1.6) {};
    \node[circle, draw=white, fill=blue!60, inner sep=2.5pt] at (axis cs:4.3,1.9) {};
    \node[circle, draw=white, fill=blue!60, inner sep=2.5pt] at (axis cs:1.9,1.9) {};

    \node[circle, draw=white, fill=red!60, inner sep=2.5pt] at (axis cs:1, 0.2) {};
    \node[circle, draw=white, fill=red!60, inner sep=2.5pt] at (axis cs:0.4, 1.15) {};
    \node[circle, draw=white, fill=red!60, inner sep=2.5pt] at (axis cs:3.3, 1) {};
    \node[circle, draw=white, fill=red!60, inner sep=2.5pt] at (axis cs:4.6, .7) {};
    \node[circle, draw=white, fill=red!60, inner sep=2.5pt] at (axis cs:2.6, 3.7) {};
    \node[circle, draw=white, fill=red!60, inner sep=2.5pt] at (axis cs:3.1, 4.7) {};
    \node[circle, draw=white, fill=red!60, inner sep=2.5pt] at (axis cs:1.1, 4.4) {};
    \node[circle, draw=white, fill=red!60, inner sep=2.5pt] at (axis cs:2.25, 4.2) {};
    \node[circle, draw=white, fill=red!60, inner sep=2.5pt] at (axis cs:2.25, 4.2) {};
    \node[circle, draw=white, fill=red!60, inner sep=2.5pt] at (axis cs:0.75, 2.2) {};
    \node[circle, draw=white, fill=red!60, inner sep=2.5pt] at (axis cs:1.5, 3) {};
    \node[circle, draw=white, fill=red!60, inner sep=2.5pt] at (axis cs:.62, 3.4) {};
    \node[circle, draw=white, fill=red!60, inner sep=2.5pt] at (axis cs:2.4, .7) {};

    \end{axis}
\end{tikzpicture}
    \caption{}
\end{subfigure}
\caption{Two examples of multivariate tree compactness. In (a) and (b) you need 10 univariate vs 1 multivariate decision(s). In (c) and (d) you need 7 univariate vs 2 multivariate decisions.}\label{fig:uni_v_multi}
\end{figure}

Researchers and industry professionals have employed decision trees in various applications including decision making in management science~\citep{magee1964} and solving integer optimization problems in operations research~\citep{land1960} since the 1960s. Due to the rise of machine learning around 1980,~\citet{breiman1984} applied decision trees to classification and regression problems. Binary decision trees are employed in a wide range of applications, including but not limited to healthcare~\citep{yoo2020,li2021}, cyber-security~\citep{maturana2011, kumar2013}, financial analysis~\citep{charlot2014, manogna2021}, and more recently fair decision making~\citep{ntoutsi2019, valdivia2021}. Further, binary decision trees are one of the most interpretable supervised machine learning methods due to their lack of a \emph{black box} nature and easy to understand branching rules and structure.~\citet{hyafil1976} show building an optimal decision tree is NP-hard and heuristic algorithms were first proposed to find approximations of decision trees as computer technology was not advanced enough to efficiently solve exact algorithms in the 1980s. Recently, optimization solvers such as Gurobi and CPLEX, have become substantially more powerful (speedup factor of 450 billion over a 20 year period) through hardware advancements, effective use of cutting plane theory, disjunctive programming for branching rules, and improved heuristic methods, as detailed by~\citet{bixby2012}. These advancements eliminate the impracticality of Mixed Integer Linear Optimization ($\milo$) formulations to solve NP-hard problems and the prejudice relevant during the inception of exact algorithms for the optimal decision tree problem, as noted by~\citet{bertsimas2017}. A majority of decision tree algorithms are for univariate decision trees (UDTs) where branching vertices test against a single training set feature; branching vertices can be thought of as axis-aligned hyperplanes. Multivariate decision trees employ branching vertices which act as separating hyperplanes by testing against sets of features. Multivariate branching rules are less interpretable, however they are more flexible than their univariate counterparts, resulting in more compact decision trees. Figure~\ref{fig:uni_v_multi} illustrates the relationship described between univariate and multivariate trees.
\vspace{-1em}

\textbf{Our Contribution}: In this paper, we focus on multivariate binary classification decision trees: trees in which each parent has exactly two children, branching vertices act as separating hyperplanes and terminal vertices assign classes. We propose two $\milo$ formulations for finding optimal binary decision trees and show their strong linear optimization (LO) relaxations compared to current $\milo$ formulations in the literature. Through experimental testing on 14 publicly available datasets, we highlight the practical application of the proposed $\milo$ formulations, their ability to scale, and strong performance against traditional branch and bound methods. Our models improve upon those currently found in the literature by taking a bi-objective approach, generating trees that are imbalanced, improve solution time through on-the-fly connectivity constraints, and we extend the current use of such shattering inequalities for decision trees by considering imbalanced decision trees. In Section~\ref{relatedwork} we review related work on binary decision trees. Further, this work presented is an extension of~\citet{alston2023} in which we extend their cut-based $\milo$ formulations for univariate decision trees to the multivariate regime. In Section~\ref{milosection}, we propose our two cut-based $\milo$ formulations ($\mcutone$ and $\mcuttwo$) for finding optimal binary trees. In Section~\ref{subprocesses}, we provide provide speedup processes for our models which have an exponential number of constraints. 
In Section~\ref{compexpr}, we provide computational experiments supporting our theoretical results and report in-sample optimization performance, out-of-sample test performance, efficiently generated Pareto frontiers for an understanding of the relationship between tree topology and out-of-sample test performance, and variations on our proposed cut-based $\milo$ models for speeding up solution time. Our goal is to provide those interested in finding optimal multivariate binary decision trees a set of implementable and flexible $\milo$ formulations.

\section{Related Work}\label{relatedwork}
Various mathematical optimization techniques have been applied to solve the binary decision tree problem. These techniques range from heuristic methods such as CART~\citep{breiman1984}, and C4.5~\citep{quinlan1993} to state of the art gradient descent methods.~\citet{murthy1994} employ hill-climbing techniques paired with randomization.~\citet{orsenigo2003} use discrete SVM operators counting misclassified points rather than measuring distance at each node of the tree; sequential LP-based heuristics are then employed to find the complete tree.~\citet{menze2011} extend oblique random forests with linear discriminate analysis (LDA) to find optimal internal splits.~\citet{wang2015} apply logistic regression to find branching hyperplanes while maintaining sparsity through a weight vector.~\citet{ balestriero2017} uses a modified hashing neural net framework with sigmoid activation functions and independent multilayer percepetrons that are equivalent to vertices of a decision tree.~\citet{zantedeschi2020} employ stochastic descent for branching attributes, auxiliary variables for linearity, and a unique tree-structured isotonic optimization algorithm for pruning-aware decision trees. Optimal randomized classification trees (ORCT) from~\citet{blanquero2021} uses a continuous optimization method for learning trees by replacing discrete binary decisions in traditional trees with probabilistic decisions. Augmented machine learning techniques have been employed to build DTs.~\cite{ balestriero2017} uses a modified hashing neural net framework with sigmoid activation functions and independent multilayer percepetrons.~\cite{zantedeschi2020} employ stochastic descent for branching attributes, auxiliary variables for linearity, and a unique tree-structured isotonic optimization algorithm.

Researchers also apply customized dynamic programming, Boolean satisfiabiility (SAT), or constraint programming (CP) to combat searching over large spaces associated with finding optimal decision trees.~\citet{aglin2020} use two branch-and-bound approaches that cache itemsets used for cutting the search space and only including vertices not in the cache in the branch-and-bound cuts.~\citet{demirovic2022} introduce constraints on the depth and number of nodes to combat scaling issues.~\citet{mctavish2022} employ guessing strategies related to feature binarization, tree depth, and bound tightening while optimizing misclassification loss and a sparsity penalty over leaves.~\citet{mazumder2022} explore the (continuously distributed) search space through the quantiles of the features.~\citet{lin2020} use a dynamic search space through hash trees and a metric that considers the relative importance of classes.~\citet{nijssen2020} use a combination of caches, itemset mining, and boolean search implemented in a CP fashion to decompose and limit the size of the decision tree problem size.~\citet{avellaneda2020} infer solutions through an incremental, generative boolean search.~\citet{janota2020} encode paths of the tree using SAT in combination with splitting the search space based on tree topologies.~\citet{narodytska2018} use a SAT-based approach for finding the smallest-size tree.~\citet{schidler2021} use a hybrid heuristic-SAT approach to generate trees over almost arbitrarily large training datasets.

\citet{breiman1984} note continued growth of the tree is indicative of successful splits and the growth itself is a one-step optimization problem; thus objective functions related to branching rather than classification metrics are sufficient. While it was~\citet{bennett1996} who propose the first $\milo$ formulation for designing optimal multivariate decision trees, in which they fix the tree structure, the number of branching vertices and the classes of leaf vertices before solving,~\citet{bertsimas2017} emphasize building a decision tree involves discrete decisions (which vertex to split on? which variable to split with?) and discrete outcomes (is a datapoint correctly classified? which leaf does a datapoint end on?). Therefore, one should consider building optimal decision trees using $\milo$ formulations.~\citet{bertsimas2017} propose OCT which outperforms CART in accuracy.~\citet{verwer2019} propose BinOCT, a binary-linear programming model aiming to reduce the dependence of the problem size on the size of the training dataset.~\citet{gunluk2018} and~\citet{firat2020} both propose column generation approaches.~\citet{gunluk2021} formulate IP models for decision trees with categorical data.

Recently,~\citet{aghaei2022} propose a flow-based $\milo$ formulation whose LP relaxation is at least as strong as that of OCT~\citep{bertsimas2017} and BinOCT~\citep{verwer2019}. They modify the structure of a traditional decision tree into a \emph{directed acyclic graph} and use a tailored Benders' decomposition is used for large size instances.~\citet{boutilier2022} 
propose a form of packing constraints~\citep{fischetti2006} which they use to find shattering inequalities related to the hyperplanes of branching vertices.~\citet{alston2023} propose two flow-based and two cut-based $\milo$ formulations, none of which use big-M formulations or a Benders' decomposition approach; the proposed formulations have strong LP relaxations but are restricted to univariate decision trees. The cut-baesd formulations of~\citet{alston2023} are motivated by the max-flow min-cut equivalency~\citep{ford1963} in directed networks, of which decision trees are, and the cut-based inequalities are the strongest thus far in the literature surrounding the optimal univariate decision tree problem. The formulations of~\citet{aghaei2022,boutilier2022,alston2023} are the main motivations of this paper.

\section{Our Formulations}\label{milosection}
Optimal multivariate decision trees provide several improvements over univariate trees trained on large datasets. Some of these improvements include (i) reducing the size of DT and overfitting, and (ii) increasing human interpretability~\citet{bennett1996, shioda2007, bertsimas2017, brodley1995}. It should be noted that the proposed formulations of~\citet{bertsimas2017, zhu2020} produce~\emph{only} balanced trees. The formulation of~\citet{bertsimas2017} showed that warm-starting solvers are still feasible with MDTs, despite the larger solution space. They also show a MILO formulation for an MDT contains the same number of binary variables as its analogous univariate decision tree formulation.

We propose two cut-based formulations, both of which have connectivity constraints that are added on-the-fly. Further, their corresponding separation problems are solved in polynomial time; the $1,v$-path of any vertex $v \in V(G_h)$ is found in $\mathcal{O}(|V|)$~\citep{kaplan2011} as a tree itself is a directed acyclic graph. The motivation behind our cut based formulation is the $P_{1,v}$ of any vertex $v \in V$ is unique since $G_h$ is a tree. Thus any vertex $c \in V(P_{1,v})$ is a valid $1,v$-separator. Through our definition of variables $q$ and $s$ we can find $1,v$-separators for a terminal vertex of a datapoint $i \in I$ to find feasible connected paths.

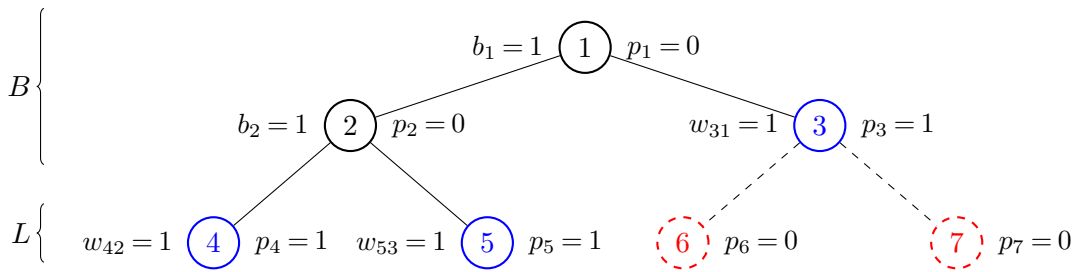
\begin{figure}[H]
\centering
\begin{tikzpicture}[scale=0.52]
\begin{scope}[every node/.style={circle,thick,draw}]
\node (1) at (7.5,8) [label={[align=center,font=\fontsize{10.5}{0}\selectfont]left:$b_1 = 1$}, label={[align=center,font=\fontsize{10.5}{0}\selectfont]right:$p_1 = 0$}]{$1$};
\node (2) at (1.5,6) [label={[align=center,font=\fontsize{10.5}{0}\selectfont]left:$b_2 = 1$}, label={[align=center,font=\fontsize{10.5}{0}\selectfont]right:$p_2=0$}] {$2$};
\node(3) at (13.5,6) [color=blue,label={[align=center,font=\fontsize{10.5}{0}\selectfont]left:$w_{31} = 1$}, label={[align=center,font=\fontsize{10.5}{0}\selectfont]right:$p_3 = 1$}] {$3$};
\node (4) at (-2,3) [color=blue, label={[align=center,font=\fontsize{10.5}{0}\selectfont]left:$w_{42} = 1$}, label={[align=center,font=\fontsize{10.5}{0}\selectfont]right:$p_4=1$}] {$4$};
\node (5) at (5,3) [color=blue,label={[align=center,font=\fontsize{10.5}{0}\selectfont]left:$w_{53} = 1$}, label={[align=center,font=\fontsize{10.5}{0}\selectfont]right:$p_5=1$}] {$5$};
\node (6) at (10,3) [color=red, dashed, label={[align=center,font=\fontsize{10.5}{0}\selectfont]right:$p_6=0$}] {$6$};
\node (7) at (17,3) [color=red, dashed, label={[align=center,font=\fontsize{10.5}{0}\selectfont]right:$p_7=0$}] {$7$};
\end{scope}
\begin{scope}[>={Stealth[black]}]
\draw [-] (1) -- (2) node [] {};
\draw [-] (1) -- (3) node [] {};
\draw [-] (2) -- (4) node [] {};
\draw [-] (2) -- (5) node [] {};
\draw [dashed] (3) -- (6) node [] {};
\draw [dashed] (3) -- (7) node [] {};

\draw[decoration={brace,mirror,raise=5pt},decorate]
  (-6,9) -- node[left=6pt] {$B$} (-6,5);
\draw[decoration={brace,mirror,raise=5pt},decorate]
  (-6,4) -- node[left=6pt] {$L$} (-6,2.5);
\end{scope}
\end{tikzpicture}\hspace{8mm}
\caption{Input decision tree $G_2=(B \cup L, E)$, branching vertex set $B=\{1,2,3\}$ and leaf vertex set $L=\{4,5,6,7\}$. Here, vertices $1$ and $2$ are assigned branching hyperplanes; vertices $3$, $4$, and $5$ are assigned to a classes 1, 2, and 3, respectively; and vertices 6 and 7 are pruned. Figure taken from~\cite{alston2023}.}
\label{fig:basetree}
\end{figure}

Given a training dataset $\mathcal{T} := \{x^i,y^i\}_{i \in I}$ consisting of datapoints indexed in the set $I$. Each row $i \in I$ of $\mathcal{T}$ consists of features, indexed in the set $F$ and collected in the vector $x^i \in [0,1]^{|F|}$, and a label $y^i$, drawn from the finite set of $K$ classes. Graph $G_h=(V,E)$ denotes the input decision tree with depth $h$, where $1 \le h \in \mathbb{N}$ is the maximal depth of a classification vertex in the assigned decision tree. The number of vertices and edges of $G_h$ are represented by $n := |V| = 2^{h+1} - 1$ and $m := |E| = 2^{h+1} - 2$, respectively. The vertex set $V$ is the union of the branching vertex set, $B \subset V$, and the leaf vertex set, $L \subset V$, with $B \cap L = \emptyset$. Figure~\ref{fig:basetree} illustrates a depth $h=2$ tree with our decision variables.
\vspace{-1em}

\subsection{Cut Based Path Feasibility}
An optimal multivariate binary classification tree can be obtained by solving a biobjective optimization problem that seeks to (i) maximize the number of correctly classified datapoints and (ii) minimize the number of branching vertices. For every vertex $v \in V$, let $P_{1,v}$ and $V(P_{1,v})$ denote the unique $1,v$-path from vertex 1 to vertex $v$ and its corresponding vertex set (including vertices $1$ and $v$), respectively. For every vertex $v \in B$, binary variable $b_v$ equals one if vertex $v$ is assigned as a branching vertex. For every vertex $v \in V$ and every class $k \in K$, binary variable $w_{vk}$ equals one if vertex $v$ is assigned to class $k$. For every vertex $v \in V$, binary variable $p_v$ equals one if a prediction class is assigned to vertex $v$. For every datapoint $i \in I$ and every vertex $v \in V$, binary variable $s^i_v$ equals one if datapoint $i$ is correctly classified at vertex $v$. For every datapoint $i \in I$ and every vertex $v \in V$, binary variable $q^i_v$ equals one if datapoint $i$ reaches vertex $v \in V$. Lastly, for every vertex $v \in B$, decision variables $(a_v, c_v) \in \mathbb{R}^{|F|\times 1}$ represents the hyperplane used at $v,~a_v^\top x^i - 1 = c_v$.
\vspace{-2em}
\begin{subequations}
\label{mCUT1model}
\begin{align}
&\max~\sum_{i \in I} \sum_{v \in V} s^i_v \label{basemax}\\
&\min~\sum_{v \in B} b_v \label{basemin}\\
& p_v = \sum_{k \in K} w_{vk} & \forall v \in V \label{base1}\\
& b_v + \sum_{u \in V(P_{1,v})} p_u =1 & \forall v \in V \label{base2}\\
& b_v = 0 & \forall v \in L\label{base3}\\
& s^i_v \leq w_{vk=y^i} & \forall k \in K,~\forall i \in I,~\forall v \in V \label{base4}\\
(\mcutone)~~~& \sum_{v \in V} s^i_v \le 1 & \forall i \in I\label{base5}\\
& q^i_{l(v)} \leq b_v & \forall v \in V\setminus \{1\},~\forall i \in I \label{base6}\\
& s^i_v \le  q^i_c & \forall v \in V \setminus \{1\},~\forall c \in V(P_v),~\forall i \in I \label{cut11}\\
& (a_v, c_v) \in \mathcal{B}_v(q) &\forall v \in B\label{branch1}\\
& b \in \{0,1\}^{|V|},~w \in \{0,1\}^{|V| \times |K|},~p \in [0,1]^{|V|}, \nonumber \\
& q \in \{0,1\}^{|I| \times |V|},~s \in \{0,1\}^{|I| \times |V|} \nonumber \\
& a \in \mathbb{R}^{|V|\times|F|},~c \in \mathbb{R}^{|V|} \label{vartype}
\end{align}
\vspace{-2em}
\end{subequations}
where,
\vspace{-1.25em}
\begin{subequations}
\begin{align}\label{hyperplane}
\mathcal{B}_v(q) = (a_v,c_v) \in \mathbb{R}^{|F|} \times \mathbb{R}:
\begin{cases}
a_v^\top x^i - 1 \leq c_v~~~\forall i \in I:q^i_{l(v)}=1\\
a_v^\top x^i + 1 \leq c_v~~~\forall i \in I:q^i_{r(v)}=1
\end{cases}
\end{align}
\end{subequations}

Here, objective function~\eqref{basemax} maximizes the number of correct classifications and objective function~\eqref{basemin} minimizes the number of branching vertices. Constraints~\eqref{base1} imply that a vertex is labeled with a prediction class if and only if it is assigned to a class $k \in K$. Constraints~\eqref{base2} imply that every vertex $v \in V$ is either assigned as a branching vertex or a vertex on the $1,v$-path is assigned to a prediction class. Constraints~\eqref{base3} imply that no leaf vertex is assigned as a branching vertex. Constraints~\eqref{base4} imply that if datapoint $i \in I$ is classified at vertex $v \in V$, then vertex $v$ is assigned to the class for which $k = y^i$. Constraints~\eqref{base5} imply that each datapoint $i \in I$ can be correctly classified in at most one vertex. Constraints~\eqref{base6} send all observations to the right child of $v$ when $v$ is not assigned as a branching vertex. Constraints~\eqref{cut11} imply that if a datapoint $i \in I$ is classified at vertex $v \in V \setminus \{1\}$, then all vertices on the path from $1$ to $v$ must be selected.

We propose another cut-based formulation whose linear optimization relaxation is stronger than that of formulation $\mcutone$ by redefining a $1,v$-separator as any vertex that separates terminal vertex $v$ or any one of its children $(\child(v))$. For every vertex $v \in V \setminus \{1\}$, we define
\begin{align*}
    \child(v) \coloneqq \{u \in V \setminus \{v\} : u>v,~\dist_{G_h}(v,u) < \infty \},
\end{align*}
where $\dist_{G_h}(v,u)$ denotes the distance between vertices $v$ and $u$ in directed graph $G_h$. By redefining the set of $1,v$-separators of a terminal vertex $v \in V$ we provide stronger lower bounds on decision variables $q$ when adding cuts at points in the branch and bound tree. This holds as a datapoint $i \in I$ must pass through $v$ to select $v$ or any one of its children as its terminal vertex. Our second proposed $\milo$ formulation for multivariate decision trees is as follows,
\begin{subequations}
\label{mCUT2model}
\begin{align}
&\max~\sum_{i \in I} \sum_{v \in V} s^i_v \label{basemax2}\\
&\min~\sum_{v \in B} b_v \label{basemin2}\\
& \eqref{base1} -\eqref{base6}~\&~\eqref{branch1} \\
(\mcuttwo)~~~& s^i_v + \sum_{u \in \child(v)} s^i_u\le  q^i_c &\forall c \in V(P_v),~\forall v \in V \setminus \{1\},~\forall i \in I \label{cut21}\\
& b \in \{0,1\}^{|V|},~w \in \{0,1\}^{|V| \times |K|},~p \in [0,1]^{|V|}, \nonumber \\
& q \in \{0,1\}^{|I| \times |V|},~s \in \{0,1\}^{|I| \times |V|} \nonumber \\
& a \in \mathbb{R}^{|V|\times|F|},~c \in \mathbb{R}^{|V|} \label{vartype2}
\end{align}
\end{subequations}
Here, constraints~\eqref{cut21} imply that if a datapoint $i \in I$ is correctly classified at vertex $v$ or one of its descendants, then the datapoint selects every vertex on the path from $1$ to $v$ excluding 1.

Our cut constraints~\eqref{cut11} and~\eqref{cut21} are exponential in nature yielding models requiring long run times when $G_h$ is assumed to be large. To combat this we introduce the constraints on-the-fly at integral or fractional points in the branch and bound tree. At fractional points we use a number of variations outline in Section~\ref{subprocesses}. We would like to emphasize the formulation can be given to solvers as a full model with constraints~\eqref{cut11} and~\eqref{cut21} presented all upfront. Observe the use of cut-based inequalities in constraints~\eqref{cut11} and~\eqref{cut21} and the biobjective, pruning aware approach in the  our models are analogous to the univariate formulations of~\cite{alston2023}.

One can think of our formulations in two main ways: i) we extend the cut-based models of~\citet{alston2023} into the multivariate regime, ii) a pruning aware, cut-based analog of~\citet{boutilier2022}. Further, similar to~\citet{alston2023} there exists a common base polytope in constraints~\eqref{base1}~\textemdash~\eqref{base6}.

\subsection{Shattering Inequalities}\label{shattering}
Formulations~\ref{mCUT1model} and~\ref{mCUT2model} aim to not use big-M constraints to define the hyperplanes of branching vertices. Instead through decision variables $q$, which track a datapoint's path through the tree, we find shattering inequalities of the form,
\vspace{-.5em}
\begin{align}\label{shatteringineq}
    & \sum_{i \in \mathcal{I}:\lambda_i=-1} q^i_{l(v)} + \sum_{i \in \mathcal{I}:\lambda_i=+1} q^i_{r(v)} \leq |\mathcal{I}|-1 &~\forall v \in B,~\forall \mathcal{I} \in I,~\lambda \in \Lambda(\mathcal{I}).
\end{align}
Here, $\lambda$ is some $\{-1,1\}$ binary classifier of $\mathcal{I}$ and $\Lambda(\mathcal{I})$ is the set of all binary classifiers of $\mathcal{I}$. The inequalities~\ref{shatteringineq} impose at least one observation at a branching vertex is not routed to the children as defined by the binary classifier used at $v$. They also hold the form of packing constraints~\citet{Cornuejols01} and were proposed for use in $\milo$ formulations of decision trees by~\citet{boutilier2022}. The motivation behind the inequalities is as follows. Let $\mathcal{A}$ be a family of binary classifiers in $\mathbb{R}^{|F|}$. Some set of observations is shattered by $\mathcal{A}$ if, for any assignment of binary labels to these observations, there exists a classifier in $\mathcal{A}$ that perfectly separates all the observations. Further, the maximum number of observations that can be shattered by $\mathcal{A}$ is the Vapnik-Chervonenkis ($VC$,~\citep{vapnik1998}) dimension of $\mathcal{A}$. If we consider $\mathcal{A}$ to be $\mathcal{B}_v(q)$ at some branching vertex $v$, then $VC(\mathcal{B}_v(q)) = |F|+1$. Further if there is some minimal set of observations in $\mathbb{R}^{|F|}$ that cannot be shattered by $\mathcal{B}_v(q)$, call it $\mathcal{C}$, then $|\mathcal{C}| \leq |F|+2$. As noted by~\citet{boutilier2022}, when $|F| \ll |I|$ the inequalities are sparse. Finding the shattering inequalities can be done by finding a minimal infeasible subsystem, MIS, (also known as irreducible infeasible system in the literature) of~\eqref{hyperplane}. We find such MISs through the operative approach of~\citet{fischetti2006}.

For the OPERATIVE($B_v(q)$) define $L_v(I) \coloneqq \{i \in I: q^i_l(v) = 1\}$ and $R_v(I) \coloneqq \{i \in I: q^i_r(v) = 1\}$. We wish to find some $I' \subseteq L_v(I),~R_v(I)$ such that $I' \cap L_v(I),~I' \cap R_v(I)$ is not perfectly separated by $(a_v,c_v)$. This is done by checking the feasibility of the dual of OPERATIVE($B_v(q)$) defined as,
\vspace{-.5em}
\begin{subequations}
\begin{align}
    \mathcal{P}_v~\coloneqq~\min &\sum_{i \in L_v(I)} w_i\lambda_i + \sum_{i \in R_v(I)} w_i\lambda_i\\
    \text{s.t.} & \sum_{i \in L_v(I)} x^i\lambda_i = \sum_{i \in R_v(I)} x^i\lambda_i\\
    & \sum_{i \in L_v(I)} \lambda_i = \sum_{i \in R_v(I)} \lambda_i = 1.
\end{align}
\end{subequations}\label{operative}
Here $w_i$ are some arbitrarily chosen weights (for example, counting the number of times a datapoint appears in a shattering inequality as proposed by~\citet{boutilier2022}). Finding support of the shattering inequalities in $\mathcal{P}_v$ is well established in the literature, e.g.~\citet{gleeson1990, amaldi2003, katerinochkina2018}. It is important to note that taking an operative approach to generate the shattering inequalities is a form of combinatorial Benders' cuts~\citet{fischetti2006} and~\citet{zanette2010} and the infeasible subsystems found have a one-to-one correspondence to the extreme points of $\mathcal{P}_v$. Further, taking a shattering inequality approach to generating linear multivariate splits is efficient but can also be applied to other binary classifiers; the separating problem becomes more difficult if splits are nonlinear.

Using the operative approach of~\citet{fischetti2006} is a natural fit for $\milo$ trained multivariate trees. Hyperplanes represent linear combinations of features perfectly separating data. It is quite clear how $\mathcal{P}_v$ is empty only if there does not exist a point $i \in I$ that is in both the convex hull of $L_v(I)$ and the convex hull of $R_v(I)$. Thus when $\mathcal{P}_v\neq\emptyset$ then $\lambda \in \Lambda(I)$ routes at least one $i \in I$ incorrectly. The shattering inequalities~\ref{shatteringineq} are thus violated by $\lambda$ and need to be introduced to the formulation. In Section~\ref{subprocesses} we detail how the shattering inequalities are added.

\section{Sub-processes}\label{subprocesses}
\begin{figure}[H]
\centering
\includegraphics[width=.99\linewidth]{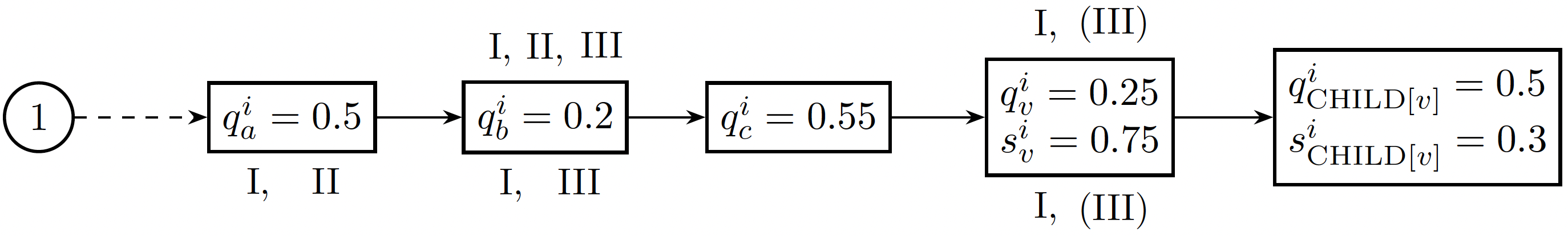}
\vspace{-.5em}\caption{Let $a, b, c$, and $v$ be nodes selected on the $1,v$-path of datapoint $i \in I$ at a fractional point in the branch and bound tree with $s^i_v$ and $q^i_u$ for $u \in P_v$ as defined. The 3 types of fractional separation cuts are indicated above/below for $\mcutone$/$\mcuttwo$, respectively. The III in parentheses is a a most violating cut considered but not added. Figure taken from~\cite{alston2023}.}\label{fig:fractypes}
\end{figure}

\textbf{Introduction of Cut-based Constraints} Our formulations $\mcutone$ and $\mcuttwo$ use a number of variations on implementing cut constraints~\eqref{cut11} and~\eqref{cut21} due to their exponential scale. We introduce integral cut constraints that cut off the relaxation solution at the root node and all integral cut constraints up front. Then we consider the cut constraints at fractional points in the branch and bound tree with three variations. The first type (I) adds all violating cuts for a datapoint in the $1,v$-path of a terminal vertex $v$; the second type (II) adds the first found violating cut in the $1,v$-path; the third type (III) adds the most violating cut, closest to the root of $G_h$, in the $1,v$-path. We consider a ``heavy'' set of user cuts (all violating cuts) and a ``light'' sets of user cuts (first found and most violating) due to~\citet{fischetti2017} who note adding too many fractional cuts may slow down solution time of $\milo$ formulations for the Steiner tree problem, which is highly related to decision trees. Figure~\ref{fig:fractypes} illustrates our variations. Such a process is also used by~\citet{alston2023} in their univariate cut-based formulations.
\vspace{-1.5em}
\vspace{-1em}
\textbf{Defining Hyperplanes} Many use the SVM problem~\citep{burges1999} to generate the optimal branching hyperplanes $(a_v,c_v)$ in~\eqref{hyperplane}. We solve the lagrangian dual of the soft-margin SVM using an $\milo$ formulation at each assigned branching vertex, $b_v=1$ from a solution of $(b,w,p,q,s)$ of $\mcutone$ or $\mcuttwo$. As mentioned earlier by taking the shattering inequalities approach of~\citet{boutilier2022} we must use a two step process to fully define the hyperplanes of vertices that have been assigned branching from solutions of our models~$\mcutone$ and~$\mcuttwo$. We describe the process in Algorithm~\ref{alg:hyperplane}.
\vspace{-.5em}
\noindent\rule{\textwidth}{.75pt}\vspace{-.75em}
\begin{algorithm}[H]
\caption{MDT x SVM}\label{alg:hyperplane}
\begin{algorithmic}
\Function{MDT\_x\_SVM}{$(b^*, w^*, p^*, q^*, s^*) \in \mcutone$ or $\mcuttwo$}
    \State $branching\_vertices = \{v;~b^*_v=1,~p^*_v=0,~w^*_{vk}=0~\forall k \in K\}$
    \For{$v \in branching\_vertices$}
        \State $L_v(I) \coloneqq \{i\in I:q^{*i}_{l(v)}=1\},~~~R_v(I) \coloneqq \{i\in I:q^{*i}_{r(v)}=1\}$
        \If{$|L_v(I)|=0$}
            \State $(a_v,c_v) = (\textbf{0}^{|F|},-1)$
        \ElsIf{$|R_v(I)|=0$}
            \State $(a_v,c_v) = (\textbf{0}^{|F|},+1)$
        \EndIf
        \State $SVM_v(I) = \delta^i~\forall i \in B_v(I)$, where
    \State ~~~~~~~~~~~~~~~~~$\{\delta^i=-1:i \in L_v(I),~~~\delta^i=+1:i \in R_v(I),~~~B_v(I)\coloneqq L_v(I)\cup R_v(I)\}$
        \State $(a_v,c_v) = SM\_SVM(SVM_v(I))$
    \EndFor
\EndFunction
\end{algorithmic}\vspace{-1.5em}
\noindent\rule{\textwidth}{.5pt}\vspace{-.5em}
$SM\_SVM(\cdot)$ is the~$\milo$ formulation of the Lagrangian dual of the soft-margin SVM.
\vspace{-.5em}
\begin{subequations}
\begin{align*}
\max_{\beta,\xi,a}~& \sum_{i \in B_v(I)} \beta_i - \frac{1}{2}\sum_{f\in F} a_f*a_f\\
& a_f = \sum_{i \in B_v(I)} \beta_i\delta^ix^i_f & \forall f \in F\\
(SM\_SVM)~~~~~~~~~& \sum_{i \in B_v(I)} \beta_i\delta^i = 0\\
& \beta \in \mathbb{R}_+^{|I|},~a \in \mathbb{R}^{|F|},~\xi \in \mathbb{R}_+^{|B_v(I)|} \nonumber.
\end{align*}\label{MIPSVM}
\end{subequations}\vspace{-2em}
\end{algorithm}\vspace{-2.75em}
\noindent\rule{\textwidth}{.75pt}
Decision variables $c = y^k-\sum_{f \in F} w_fx^k_f$, where
\vspace{-.75em}
\begin{align*}
k=\argmin_i\{\alpha_i > 0\}, \text{and}~y^i \coloneqq \{+1~\forall i: q^i_{r(v)},~-1~\forall i: q^i_{l(v)}\}
\end{align*}
We choose to use the soft-margin SVM for a number of reasons. Finding the branching hyperplanes of~\eqref{hyperplane} by solving the hard margin linear SVM problem needs the data to be disjoint, which is traditionally the case when given low dimensional training data. As the size of the dataset increases, often subsets $L_v(I), R_v(I)$ will intersect in many dimensions of $F$ yielding hard margin SVM algorithms invalid. Further as we implement our $\milo$ models with a time limit, all shattering inequalities~\eqref{shatteringineq} may not be found resulting in data that is not perfectly separated. When the soft-margin SVM is infeasible for a given assigment of $y$, we use generic hyperplanes for~\eqref{hyperplane}. This may lead to weak separation of the data at said branching vertex, and when done sequentially global tree classification rates tend to suffer.

\textbf{Generating Shattering Inequalities} For finding shattering inequalities~\eqref{shatteringineq} we use a decomposition approach involving a master $\milo$ problem and an LP feasibility subproblem. Rather than solve $\mcutone$ or $\mcuttwo$ entirely, we remove constraints~\eqref{branch1}, leaving a master problem involving only decision variables $(b,w,p,q,s)$. Our LP feasibility subproblem only involves decision variables $(a_v,c_v)~\forall v \in B$. Given that if $\mathcal{P}_v=\emptyset$ then $\lambda$ perfectly separates $\mathcal{I} \in I$, our goal is simply to check the feasibility of $\mathcal{P}_v$ at each $v \in B$ for a solution of $(b,w,p,q,s)$ generated by our master problem. At integral points in the branch and bound tree of our master problem we generate sets $L_v(I),~R_v(I)$ for each $v \in B$ from solutions of $q$. We then pass sets $L_v(I), R_v(I)$ to $\mathcal{P}_v$. If $\mathcal{P}_v =\emptyset~\forall v\in B$, then the solution of $q$ is valid for all $v \in B$. If not, we add the corresponding inequalities~\eqref{shatteringineq} to the master problem. Further by updating objective weights $w$ of $\mathcal{P}_v$ we can generate multiple inequalities at once by finding multiple extreme points of $\mathcal{P}_v$, also noted by~\citet{boutilier2022}.

Our process for generating inequalities~\eqref{shatteringineq} parallels that of~\cite{boutilier2022}, with fundamental differences. We both check for MIS subsystems at integral nodes in the branch-and-bound tree, form LP feasibility subproblems from the left-right exit direction of entering datapoints, and add inequalities inspired by~\cite{fischetti2006}. In~\citet{boutilier2022} they produce balanced trees and thus all final branching vertices are known \emph{a-priori} (original branching set $B$), while in our model we allow for pruning. This difference in final tree topology yields variations at which vertices of the decision tree we check for MIS subsystems. In our model we only check at nodes where $b_v=1,~p_v=0$ and $w_{vk}=0~\forall k\in K$. In~\citet{boutilier2022} they check at all $v \in B$. In our model sets $R_v(I), L_v(I)$ (corresponding to the support of the MIS subsystems) are determined by values of decision variables $q$ whereas in~\citet{boutilier2022} such sets are determined by the values of their flow-based decision variables.

The one-to-one correspondence between the support of~$\mathcal{P}_v$ and the MIS of $B_v(q)$ does not guarantee redundant cuts will not be generated. For different valid integeral solutions of $(p,b,w,q,s)$ it may be that values of $b_v$ for some $v\in V$ are shared. Our process for generating shattering inequalities~\eqref{shatteringineq} (and that of~\citet{boutilier2022}) we would generate the corresponding cut of the MIS at $b_v$ more than once while parsing through the integral solutions of the branch-and-bound tree. Such repeated cut generation is evident by the reduced accuracy of the shattering inequalities approach vs the traditional big-M approach, later discussed in our computational experiments.

\section{Computational Experiments}\label{compexpr}
In this Section we provide experiments on publicly available datasets to benchmark our proposed formulations, $\mcutone$ and~$\mcuttwo$, against four methods from the literature: two~$\milo$ approach,~$\soct$~\citet{boutilier2022} and $\octh$~\citet{bertsimas2017}; a branch-and-bound approach, DL8.5~\citet{aglin2020} and the industry standard heuristic,~$\cart$~\citet{breiman1984}. Note that models~$\soct$ and~$\octh$ only produce balanced decision trees.

\subsection{Experimental Setup}
We run all experiments on an Intel(R) Core(TM) i7-9800X CPU (3.8Ghz, 19.25MB, 165W) using 1 core and 16GB RAM. Code is written in Python 3.9.~$\milo$ formulations are solved using Gurobi 10.0. All models have a 15 minute time limit. Code is available at \url{https://github.com/brandalston/MDT}. We use 6 categorical and 8 numerical datasets from the UCI ML repository (\url{http://archive.ics.uci.edu/ml/index.php}). For categorical datasets we use the standard one-hot encoding. For numerical datasets we perform simple normalization to $[0,1]$ for when we perform the Algorithm~\ref{alg:hyperplane} to determine the MDT's hyperplanes.
\vspace{-.5em}
\begin{table}[H]
\renewcommand{\arraystretch}{1.1}\setlength{\tabcolsep}{4pt}
\fontsize{8}{7}\selectfont\centering
\caption{Dataset size ($|I|$), number of encoded features ($|F|$), number of classes ($|K|$) and featureset type: categorical or numerical ($\mathcal{C}/\mathcal{N}$).}\label{tab:multi_datasets}
\begin{tabular}{c||cccccccccccccc}\hline
  Dataset & bank & blood & b.c & climate & fico & glass & image & ion & iris & monk1 & parkin & soy & spect & t.t.t. \\\hline
  $|I|$ & 1372 & 747 & 286 & 540 & 10459 & 214 & 2305 & 351 & 150 & 124 & 195 & 47  & 267 & 958 \\
  $|F|$ & 4   & 4   & 43  & 18  & 34  & 9   & 19  & 34  & 4   & 17  & 22  & 72  & 44  & 27 \\
  $|K|$ & 2   & 2   & 2   & 2   & 2   & 6   & 7   & 2   & 3   & 2   & 2   & 4   & 2   & 2 \\
  Type & $\mathcal{N}$ & $\mathcal{N}$ & $\mathcal{C}$ & $\mathcal{N}$ & $\mathcal{C}$ & $\mathcal{N}$ & $\mathcal{N}$ & $\mathcal{N}$ & $\mathcal{N}$ & $\mathcal{C}$ & $\mathcal{N}$ & $\mathcal{C}$ & $\mathcal{C}$ & $\mathcal{C}$ \\\hline
\end{tabular}
\end{table}
The~$\milo$ models inherently allow for pruning (constraints~\eqref{base2}) and the branch-and-bound models are given initial lower bounds, which aims for both classes of models to prevent over fitting in a solution. However, it is well known tuned hyperparameters related to tree sparsity are needed for maximizing out-of-sample test accuracy. Thus, we remove objective~\eqref{basemin} and modify~\eqref{basemax} to
\vspace{-.5em}
\begin{subequations}
\begin{align*}
    \max~(1-\lambda)\sum_{v\in V\setminus\{0\}}\sum_{i\in I}s^i_v-\lambda\sum_{v\in V} b_v,
    \end{align*}
\end{subequations}
where~$\lambda\in[0,1]$ is a hyperparameter used to control tree sparsity. For each dataset we create 5 random 75-25\% train-test splits and train trees of depth~$h \in \{2,3,4,5\}$. We tune~$\lambda \in\{0,0.1,\dots,0.9\}$ with~$15\%$ of the dataset, taken as subset of the training set. Each calibration model has a time limit of 15 minutes. No warm starts are used unless otherwise stated. For results related to balanced trees we add constraints of the form, $b_v = 0~~\forall v\in B$ to generate such balanced DTs.
\raggedbottom

\subsection{Experimental Results}
\vspace{-1em}
\begin{table}[H]
\fontsize{9.5}{8.25}\selectfont\centering
\caption{Average~$\pm$ standard deviation of solution time (s), in-sample optimality gap \%-age (in parentheses) if the 15 min TL was reached. Best in \textbf{bold} for MILO models.}\label{tab:multi_time_gap}
\renewcommand{\arraystretch}{1.15}
\begin{tabular}{l||cccccc}
Dataset & $\mcutone$ & $\mcuttwo$ & $\octh$ & $\soct$ & DL8.5 & CART \\\hline

    & \multicolumn{6}{c}{$h$=2}\\\hline
  bank & 540$\pm$389 & 540$\pm$389 & \textbf{327$\pm$441} & (80.68$\pm$2.67) & 0$\pm$0 & 0$\pm$0 \\
  blood & \textbf{730$\pm$0} & (51$\pm$17.39) & (100$\pm$0) & (31.1$\pm$1.64) & 0$\pm$0 & 0$\pm$0 \\
  b.c. & \textbf{2$\pm$3} & 19$\pm$24 & 894$\pm$14 & (28.24$\pm$5.52) & 0$\pm$0 & 0$\pm$0 \\
  climate & \textbf{1$\pm$0} & 2$\pm$1 & 14$\pm$18 & 732$\pm$377 & 0$\pm$0 & 0$\pm$0 \\
  fico & \textbf{724$\pm$1 }& \textbf{724$\pm$1} & (0$\pm$0) & (91.65$\pm$0.42) & 0$\pm$0 & 0$\pm$0 \\
  glass & \textbf{209$\pm$387} & 376$\pm$394 & (100$\pm$0) & (97.88$\pm$28.79) & 0$\pm$0 & 0$\pm$0 \\
  image & \textbf{361$\pm$328} & 392$\pm$467 & (100$\pm$0) & (96.52$\pm$1.01) & 0$\pm$0 & 0$\pm$0 \\
  ion & \textbf{2$\pm$1} & \textbf{2$\pm$2} & 17$\pm$14 & 710$\pm$355 & 0$\pm$0 & 0$\pm$0 \\
  iris & 4$\pm$8 & \textbf{1$\pm$2} & \textbf{1$\pm$2} & 57$\pm$118 & 0$\pm$0 & 0$\pm$0 \\
  monk1 & \textbf{0$\pm$0 }& \textbf{0$\pm$0} & 1$\pm$0 & 166$\pm$78 & 0$\pm$0 & 0$\pm$0 \\
  parkin & \textbf{0$\pm$0} & \textbf{0$\pm$0} & 28$\pm$19 & 364$\pm$489 & 0$\pm$0 & 0$\pm$0 \\
  soy & \textbf{0$\pm$0} & \textbf{0$\pm$0} & \textbf{0$\pm$0} & \textbf{0$\pm$0} & 0$\pm$0 & 0$\pm$0 \\
  spect & \textbf{36$\pm$56} & 181$\pm$399 & (83.28$\pm$11.86) & (47.33$\pm$9.33) & 0$\pm$0 & 0$\pm$0 \\
  t.t.t. & 185$\pm$400 & 185$\pm$400 & \textbf{173$\pm$115} & 747$\pm$342 & 0$\pm$0 & 0$\pm$0 \\\hline
    
    & \multicolumn{6}{c}{$h$=3}\\\hline
  bank & \textbf{541$\pm$0 }& \textbf{541$\pm$0} & 783$\pm$261 & (64.42$\pm$35.66) & 0$\pm$0 & 0$\pm$0 \\
  blood & \textbf{720$\pm$0} & \textbf{720$\pm$228} & (100$\pm$0) & (31.23$\pm$1.78) & 0$\pm$0 & 0$\pm$0 \\
  b.c. & \textbf{236$\pm$385} & 316$\pm$399 & 820$\pm$178 & 770$\pm$290 & 0$\pm$0 & 0$\pm$0 \\
  climate & 185$\pm$202 & \textbf{2$\pm$0} & 29$\pm$18 & 598$\pm$420 & 1$\pm$1 & 0$\pm$0 \\
  fico & MEM & MEM & (100$\pm$0) & \textbf{(91.65$\pm$0.42)} & 1$\pm$0 & 0$\pm$0 \\
  glass & \textbf{(64.97$\pm$29.44)} & (79.27$\pm$30.99) & (100$\pm$0) & (165.19$\pm$19.56) & 0$\pm$0 & 0$\pm$0 \\
  image & MEM & MEM & \textbf{(100$\pm$0)} & (212.37$\pm$48.21) & 1$\pm$1 & 0$\pm$0 \\
  ion & \textbf{1$\pm$0 }& \textbf{1$\pm$0} & 63$\pm$48 & 41$\pm$73 & 2$\pm$2 & 0$\pm$0 \\
  iris & \textbf{1$\pm$1} & 25$\pm$32 & 4$\pm$2 & 200$\pm$391 & 0$\pm$0 & 0$\pm$0 \\
  monk1 & \textbf{1$\pm$1} & \textbf{1$\pm$2 }& \textbf{1$\pm$1} & 23$\pm$18 & 0$\pm$0 & 0$\pm$0 \\
  parkin & \textbf{0$\pm$0} & \textbf{0$\pm$0} & \textbf{}63$\pm$104 & 720$\pm$402 & 1$\pm$1 & 0$\pm$0 \\
  soy & \textbf{0$\pm$0} & \textbf{0$\pm$0} & 1$\pm$0 & 0$\pm$0 & 0$\pm$0 & 0$\pm$0 \\
  spect & 418$\pm$364 & \textbf{301$\pm$382} & (0$\pm$12.78) & (33.14$\pm$13.99) & 0$\pm$0 & 0$\pm$0 \\
  t.t.t. & 543$\pm$396 & 371$\pm$0 & \textbf{309$\pm$353} & 721$\pm$401 & 0$\pm$0 & 0$\pm$0 \\\hline
    
    & \multicolumn{6}{c}{$h$=4}\\\hline
  bank & \textbf{373$\pm$0} & 544$\pm$0 & 573$\pm$374 & 721$\pm$401 & 0$\pm$0 & 0$\pm$0 \\
  blood & MEM & MEM & (100$\pm$0) & \textbf{(31.23$\pm$1.78)} & 0$\pm$0 & 0$\pm$0 \\
  b.c. & 578$\pm$0 &\textbf{ 541$\pm$0} & (100$\pm$0) & (20.25$\pm$4.42) & 2$\pm$0 & 0$\pm$0 \\
  climate & 364$\pm$347 & 362$\pm$327 & \textbf{84$\pm$28} & 564$\pm$462 & 22$\pm$23 & 0$\pm$0 \\
  fico & MEM & MEM & (100$\pm$0) & \textbf{(91.65$\pm$0.42)} & 11$\pm$0 & 0$\pm$0 \\
  glass & \textbf{(0$\pm$29.44)} & MEM & \textbf{(0$\pm$0)} & (173.01$\pm$24.15) & 1$\pm$1 & 0$\pm$0 \\
  image & MEM & MEM & \textbf{(100$\pm$0)} & (365.63$\pm$180.53) & 22$\pm$24 & 0$\pm$0 \\
  ion & 128$\pm$263 & \textbf{25$\pm$31} & 50$\pm$17 & 585$\pm$431 & 78$\pm$82 & 0$\pm$0 \\
  iris & 141$\pm$127 & \textbf{82$\pm$81} & 33$\pm$36 & 188$\pm$398 & 0$\pm$0 & 0$\pm$0 \\
  monk1 & \textbf{1$\pm$2 }& 9$\pm$8 & 2$\pm$1 & 37$\pm$44 & 0$\pm$0 & 0$\pm$0 \\
  parkin & \textbf{1$\pm$0} & \textbf{1$\pm$0 }& 56$\pm$51 & 602$\pm$410 & 13$\pm$15 & 0$\pm$0 \\
  soy & \textbf{0$\pm$0} &\textbf{ 0$\pm$0} & 1$\pm$1 & 0$\pm$0 & 0$\pm$0 & 0$\pm$0 \\
  spect & \textbf{460$\pm$0} & 575$\pm$0 & (100$\pm$0) & (32.37$\pm$11.32) & 9$\pm$0 & 0$\pm$0 \\
  t.t.t. & 722$\pm$0 & \textbf{370$\pm$0} & 469$\pm$282 & (54.53$\pm$3.32) & 1$\pm$0 & 0$\pm$0 \\\hline
    
    & \multicolumn{6}{c}{$h$=5}\\\hline
  bank & MEM & \textbf{366$\pm$0} & 805$\pm$214 & 543$\pm$489 & 0$\pm$0 & 0$\pm$0 \\
  blood & MEM & MEM & (100$\pm$0) & \textbf{(31.23$\pm$1.78)} & 0$\pm$0 & 0$\pm$0 \\
  b.c. & MEM & MEM & (100$\pm$0) & \textbf{(7.48$\pm$8.3)} & 25$\pm$2 & 0$\pm$0 \\
  climate & 605$\pm$0 & 832$\pm$0 & \textbf{88$\pm$50} & (8.59$\pm$0.97) & 209$\pm$221 & 0$\pm$0 \\
  fico & MEM & MEM & (100$\pm$0) & \textbf{(91.65$\pm$0.42)} & 138$\pm$3 & 0$\pm$0 \\
  glass & MEM & MEM & \textbf{(100$\pm$0)} & (177.56$\pm$16.05) & 4$\pm$4 & 0$\pm$0 \\
  image & MEM & MEM & \textbf{(100$\pm$0)} & (563.1$\pm$3.41) & 445$\pm$469 & 0$\pm$0 \\
  ion & 182$\pm$379 & \textbf{4$\pm$3} & 164$\pm$82 & 393$\pm$467 & 232$\pm$314 & 0$\pm$0 \\
  iris & 229$\pm$317 & 321$\pm$333 & \textbf{19$\pm$14} & 566$\pm$460 & 0$\pm$0 & 0$\pm$0 \\
  monk1 & \textbf{2$\pm$2} & 14$\pm$17 & 5$\pm$6 & 41$\pm$31 & 0$\pm$0 & 0$\pm$0 \\
  parkin & 21$\pm$0 &\textbf{ 2$\pm$0} &\textbf{} 63$\pm$35 & 301$\pm$214 & 10$\pm$14 & 0$\pm$0 \\
  soy & \textbf{0$\pm$0} & \textbf{0$\pm$0} & 2$\pm$0 & \textbf{0$\pm$0} & 0$\pm$0 & 0$\pm$0 \\
  spect & MEM & MEM & (100$\pm$0) & \textbf{(22.95$\pm$16.26)} & 100$\pm$7 & 0$\pm$0 \\
  t.t.t. & 729$\pm$0 & 214$\pm$0 & 717$\pm$261 & \textbf{33$\pm$33} & 8$\pm$0 & 0$\pm$0 \\\hline
  \end{tabular}
\end{table}

\begin{table}[H]
\fontsize{10}{9}\selectfont\centering
\caption{Average~$\pm$ standard deviation of out-of-sample accuracy (\%). Best in \textbf{bold}.}\label{tab:multi_accuracy}
\renewcommand{\arraystretch}{1.2}
\begin{tabular}{l||cccccc}
Dataset & $\mcutone$ & $\mcuttwo$ & $\octh$ & $\soct$ & DL8.5 & CART \\\hline

    & \multicolumn{6}{c}{$h$=2}\\\hline
  bank & 60.47$\pm$6.84 & 60.47$\pm$6.84 & \textbf{99.59$\pm$0.49} & 56.09$\pm$2.44 & 70.23$\pm$16.93 & 86.41$\pm$1.42 \\
  blood & 72.73$\pm$9.27 & 76.36$\pm$3.03 & \textbf{77.75$\pm$4.24} & 76.26$\pm$2.74 & 76.2$\pm$2.92 & 76.58$\pm$3.08 \\
  b.c. & 51.39$\pm$14.04 & 52.22$\pm$14.04 & 62.5$\pm$6.29 & 64.17$\pm$5.33 & \textbf{65.56$\pm$3.51} & 65.56$\pm$2.48 \\
  climate & 66.96$\pm$34.15 & 84$\pm$15.3 & 91.26$\pm$1.42 & 90.81$\pm$4.7 & \textbf{93.11$\pm$2.21} & 89.63$\pm$2.46 \\
  fico & 51.52$\pm$1.88 & 50.68$\pm$2.41 & 57.93$\pm$6.01 & 52.24$\pm$0.34 & \textbf{70.83$\pm$0.86} & 68.88$\pm$0.54 \\
  glass & 43.33$\pm$4.14 & 47.04$\pm$6.23 & \textbf{52.22$\pm$7.34} & 36.3$\pm$5.34 & 41.67$\pm$17 & 42.59$\pm$6.14 \\
  image & 29.43$\pm$16.72 & 29.12$\pm$16.39 & 24.33$\pm$6.77 & 24.47$\pm$0.81 & \textbf{30.57$\pm$17.17} & 25.27$\pm$1.01 \\
  ion & 67.5$\pm$6.15 & 63.18$\pm$3.73 & \textbf{83.86$\pm$3.35} & 83.41$\pm$3.82 & 80.8$\pm$4.8 & 80.68$\pm$2.41 \\
  iris & \textbf{95.79$\pm$3.53} & 95.26$\pm$4.32 & 95.26$\pm$5.06 & 95.26$\pm$4.32 & 54.47$\pm$31.11 & 65.79$\pm$4.16 \\
  monk1 & 65.16$\pm$1.77 & 64.52$\pm$1.77 & 85.16$\pm$13.42 & \textbf{88.39$\pm$5.4} & 82.58$\pm$6.31 & 75.48$\pm$9.84 \\
  parkin & 78.57$\pm$5.51 & 74.69$\pm$7.44 & 82.04$\pm$8.46 & 75.92$\pm$3.35 & 83.06$\pm$5.36 & \textbf{85.71$\pm$5.2} \\
  soy & \textbf{100$\pm$0} & \textbf{100$\pm$0} & 91.67$\pm$8.33 & \textbf{100$\pm$0} & 96.67$\pm$4.3 & 50$\pm$13.18 \\
  spect & 52.54$\pm$7.42 & 53.73$\pm$8.04 & 65.37$\pm$6.62 & 60.9$\pm$6.1 & \textbf{71.94$\pm$3.5} & 68.66$\pm$5.38 \\
  t.t.t. & 72.75$\pm$10.73 & 72.75$\pm$10.73 & \textbf{94.58$\pm$0.83} & 72.58$\pm$13.4 & 67.08$\pm$2.2 & 71.33$\pm$2.47 \\\hline

      & \multicolumn{6}{c}{$h$=3}\\\hline
  bank & 73.53$\pm$20.51 & 73.53$\pm$5.18 & \textbf{99.13$\pm$0.74} & 64.55$\pm$19.79 & 73.35$\pm$20.22 & 90.79$\pm$0.73 \\
  blood & 76.58$\pm$2.31 & 76.79$\pm$1.99 & \textbf{77.01$\pm$2.3} & 76.58$\pm$3.08 & 75.99$\pm$3.21 & 76.58$\pm$3.08 \\
  b.c. & 61.67$\pm$7.71 & 65$\pm$4.44 & 63.89$\pm$2.41 & 64.72$\pm$3.75 & 67.22$\pm$5.29 & \textbf{67.22$\pm$3.04} \\
  climate & 90.07$\pm$2.49 & 79.56$\pm$24.3 & 91.26$\pm$2.42 & 91.41$\pm$2.94 & \textbf{91.63$\pm$2.07} & 90.81$\pm$2.85 \\
  fico & 50.68$\pm$2.41 & 50.68$\pm$2.41 & 56.56$\pm$5.04 & 52.24$\pm$0.34 & \textbf{71.14$\pm$0.79} & 68.88$\pm$0.54 \\
  glass & 51.48$\pm$3.84 & 51.48$\pm$3.56 & \textbf{52.96$\pm$6.09} & 34.07$\pm$7.48 & 45.56$\pm$19.23 & 50.37$\pm$4.97 \\
  image & 14.04$\pm$1.75 & 14.04$\pm$1.75 & 19.17$\pm$7.63 & 27.31$\pm$6.57 & 38.73$\pm$25.77 & \textbf{40$\pm$1.22} \\
  ion & 66.36$\pm$18.33 & 68.18$\pm$5.02 & 79.32$\pm$3.54 & 85.23$\pm$4.25 & 82.5$\pm$3.64 & \textbf{88.86$\pm$3.45} \\
  iris & 94.74$\pm$1.86 & \textbf{96.32$\pm$3.99} & 91.58$\pm$7.54 & 83.68$\pm$24.85 & 57.63$\pm$33.37 & 93.68$\pm$4.4 \\
  monk1 & 65.81$\pm$2.89 & 63.87$\pm$5.77 & 80$\pm$8.35 & 76.13$\pm$5.4 & \textbf{87.1$\pm$6.8} & 69.68$\pm$7.77 \\
  parkin & 78.78$\pm$3.1 & 80$\pm$2.66 & 80.82$\pm$8.24 & 79.59$\pm$4.33 & 83.27$\pm$5.34 & \textbf{85.71$\pm$5.2} \\
  soy & \textbf{98.33$\pm$3.73} & \textbf{98.33$\pm$3.73} & 75$\pm$16.67 & \textbf{98.33$\pm$3.73} & 96.67$\pm$4.3 & 68.33$\pm$12.36 \\
  spect & 55.82$\pm$5.02 & 58.51$\pm$4.65 & 62.99$\pm$6.87 & 58.51$\pm$4.27 & 65.07$\pm$4.41 & \textbf{67.76$\pm$5.23} \\
  t.t.t. & 54$\pm$18.01 & 60$\pm$18.38 & \textbf{93.75$\pm$1.98} & 85$\pm$14.98 & 72.75$\pm$1.35 & 69.67$\pm$3.31 \\\hline

      & \multicolumn{6}{c}{$h$=4}\\\hline
  bank & 79.83$\pm$5.18 & 72.54$\pm$5.18 & \textbf{99.3$\pm$0.6} & 65.36$\pm$19.47 & 75.39$\pm$22.36 & 94.23$\pm$1.19 \\
  blood & 66.84$\pm$23.19 & 66.84$\pm$23.19 & 76.26$\pm$2.58 & 76.58$\pm$3.08 & 77.22$\pm$2.31 & \textbf{77.65$\pm$2.02} \\
  b.c. & 61.11$\pm$10.64 & 62.78$\pm$10.04 & 63.61$\pm$5.14 & 64.72$\pm$3.75 & 64.44$\pm$7.21 & \textbf{67.78$\pm$2.67} \\
  climate & \textbf{91.7$\pm$2.46} & 90.67$\pm$2.74 & 89.19$\pm$4.43 & 91.26$\pm$2.84 & 90.07$\pm$3.62 & 88.89$\pm$2.22 \\
  fico & 50.68$\pm$2.41 & 50.68$\pm$2.41 & 52.24$\pm$0.34 & 52.24$\pm$0.34 & \textbf{71.27$\pm$0.89} & 70.07$\pm$0.25 \\
  glass & 44.44$\pm$14.28 & 32.59$\pm$14.56 & 43.7$\pm$12.87 & 32.22$\pm$5.17 & 46.3$\pm$20.47 & \textbf{59.63$\pm$4.42} \\
  image & 14.04$\pm$1.75 & 14.04$\pm$1.75 & 21.98$\pm$10.67 & 19.31$\pm$6.66 & 47$\pm$34.45 & \textbf{57.23$\pm$6.15} \\
  ion & 68.41$\pm$9.28 & 65.91$\pm$11.39 & 82.05$\pm$4.71 & 72.27$\pm$13.65 & 81.7$\pm$5.47 & \textbf{88.41$\pm$2.19} \\
  iris & 93.16$\pm$3.99 & 95.26$\pm$2.63 & 90$\pm$2.88 & 90$\pm$10.91 & 60.79$\pm$36.79 & \textbf{96.84$\pm$4.32} \\
  monk1 & 68.39$\pm$8.22 & 66.45$\pm$7.43 & 72.26$\pm$14.17 & 68.39$\pm$8.35 & \textbf{100$\pm$0} & 82.58$\pm$6.69 \\
  parkin & 68.16$\pm$21.08 & 74.29$\pm$5.32 & 78.78$\pm$5.32 & 78.37$\pm$4.23 & 83.47$\pm$5.13 & \textbf{87.76$\pm$5} \\
  soy & \textbf{100$\pm$0} & \textbf{100$\pm$0} & 81.67$\pm$19 & 85$\pm$10.87 & 85$\pm$19.56 & 96.67$\pm$4.56 \\
  spect & 59.7$\pm$2.71 & 60$\pm$4.55 & 61.49$\pm$4.88 & 55.82$\pm$2.26 & \textbf{66.57$\pm$4.73} & 66.27$\pm$6.03 \\
  t.t.t. & 68.17$\pm$14.59 & 76.83$\pm$15.05 & \textbf{95.42$\pm$1.98} & 67.17$\pm$4.12 & 81.5$\pm$1.41 & 73.58$\pm$1.9 \\\hline

      & \multicolumn{6}{c}{$h$=5}\\\hline
  bank & 55.74$\pm$5.18 & 71.43$\pm$5.18 & \textbf{99.3$\pm$0.6} & 73.7$\pm$23.83 & 76.21$\pm$23.21 & 95.86$\pm$1.19 \\
  blood & 66.84$\pm$23.19 & 66.84$\pm$23.19 & 76.04$\pm$1.94 & 76.58$\pm$3.08 & \textbf{77.22$\pm$2.34} & 76.47$\pm$2.11 \\
  b.c. & \textbf{66.39$\pm$3.01} & 65.28$\pm$4.71 & 61.11$\pm$8.95 & 61.11$\pm$4.39 & 65.83$\pm$5.37 & \textbf{66.39$\pm$3.32} \\
  climate & \textbf{90.67$\pm$2.43} & 90.07$\pm$2.49 & 89.78$\pm$3.07 & 89.63$\pm$2.46 & 89.04$\pm$4.08 & 90.52$\pm$2.42 \\
  fico & 50.68$\pm$2.41 & 50.68$\pm$2.41 & 52.24$\pm$0.34 & 52.24$\pm$0.34 & \textbf{71.21$\pm$1.01} & 70.74$\pm$0.99 \\
  glass & 23.33$\pm$5.3 & 32.59$\pm$19.62 & 51.11$\pm$6.23 & 31.48$\pm$5.86 & 45.19$\pm$19.8 & \textbf{62.22$\pm$6.09} \\
  image & 14.04$\pm$1.75 & 14.04$\pm$1.75 & 14.38$\pm$4.29 & 11.85$\pm$0.4 & 50.88$\pm$38.49 & \textbf{70.92$\pm$6.87} \\
  ion & 80$\pm$8.3 & 82.73$\pm$3.15 & 81.59$\pm$7.07 & 77.05$\pm$13.88 & 79.77$\pm$4.44 & \textbf{85.68$\pm$3.37} \\
  iris & 92.63$\pm$3.43 & 94.74$\pm$2.63 & 91.05$\pm$5.13 & 54.74$\pm$35.47 & 57.11$\pm$32.92 & \textbf{96.84$\pm$3.43} \\
  monk1 & 69.68$\pm$2.28 & 65.81$\pm$6.69 & 70.97$\pm$10.2 & 69.03$\pm$12.2 & \textbf{92.26$\pm$3.47} & 80$\pm$8.35 \\
  parkin & 78.78$\pm$4.47 & 78.37$\pm$4.7 & 80$\pm$1.71 & 77.55$\pm$9.89 & 83.47$\pm$4.85 & \textbf{88.57$\pm$4.91} \\
  soy & \textbf{98.33$\pm$3.73} & 96.67$\pm$4.56 & 75$\pm$11.79 & 95$\pm$7.45 & 86.67$\pm$8.96 & 96.67$\pm$4.56 \\
  spect & 57.91$\pm$4.65 & 55.22$\pm$5.32 & 58.21$\pm$4.6 & 57.91$\pm$5.32 & 60.6$\pm$4.41 & \textbf{63.58$\pm$5.64} \\
  t.t.t. & 55.67$\pm$18.87 & 89.58$\pm$14.01 & 81.42$\pm$13.78 & \textbf{94.42$\pm$1.2} & 87$\pm$2.07 & 82.92$\pm$1.84 \\\hline
  \end{tabular}
\end{table}

\begin{table}[H]
\fontsize{10}{9}\selectfont\centering
\caption{Average~$\pm$ standard deviation of in-sample accuracy (\%). Best in \textbf{bold}.}\label{tab:mutli_in_accuracy}
\renewcommand{\arraystretch}{1.1}
\begin{tabular}{l||cccccc}
Dataset & $\mcutone$ & $\mcuttwo$ & $\octh$ & $\soct$ & DL8.5 & CART \\\hline  
  & \multicolumn{6}{c}{$h$=2}\\\hline
  banknote & 59.9$\pm$5.96 & 59.9$\pm$5.96 & \textbf{99.77$\pm$0.28} & 55.35$\pm$0.81 & 71.92$\pm$16.71 & 85$\pm$0.47 \\
  blood & 74.71$\pm$6.4 & 76.64$\pm$3.34 & \textbf{77.82$\pm$1.09} & 76.29$\pm$0.95 & 76.57$\pm$0.89 & 76.21$\pm$1.03 \\
  b.c. & 58.6$\pm$21.62 & 58.6$\pm$21.88 & \textbf{95.05$\pm$1.94} & 77.94$\pm$3.24 & 78.69$\pm$1.19 & 73.64$\pm$0.71 \\
  climate & 68.94$\pm$37.6 & 87.36$\pm$16.7 & \textbf{97.48$\pm$0.98} & 96.44$\pm$2.92 & 91.75$\pm$0.83 & 92.1$\pm$0.82 \\
  fico & 0$\pm$2 & 0$\pm$2.39 & 0$\pm$6.69 & 52.18$\pm$0.11 & 0$\pm$0.29 & \textbf{69.7$\pm$0.28} \\
  glass & 59.25$\pm$8.75 & 65.13$\pm$7.77 & \textbf{66.38$\pm$8.96} & 46.25$\pm$6.99 & 49$\pm$12.81 & 47.88$\pm$1.8 \\
  image & 32.26$\pm$19.2 & 32.28$\pm$19.46 & 26.78$\pm$6.93 & 29.91$\pm$0.18 & \textbf{33.08$\pm$16.76} & 29.65$\pm$0.45 \\
  ion & 71.79$\pm$13.75 & 66.46$\pm$2.07 & 94.68$\pm$1.32 & \textbf{95.51$\pm$9.61} & 86.16$\pm$2.72 & 83.8$\pm$0.95 \\
  iris & 99.29$\pm$1.16 & 98.39$\pm$2.71 & 98.04$\pm$2.13 & \textbf{99.82$\pm$0.4} & 62.14$\pm$25.44 & 66.96$\pm$1.41 \\
  monk1 & 0$\pm$7.29 & 0$\pm$7.29 & 0$\pm$1.52 & \textbf{100$\pm$0} & 0$\pm$1.6 & 72.69$\pm$3.28 \\
  parkin & 82.05$\pm$5.76 & 76.44$\pm$2.09 & \textbf{94.79$\pm$2.14} & 94.52$\pm$7.57 & 82.33$\pm$7.57 & 86.85$\pm$1.64 \\
  soy & \textbf{100$\pm$0} & \textbf{100$\pm$0} & 96$\pm$7.45 & \textbf{100$\pm$0} & \textbf{100$\pm$0} & 60$\pm$4.52 \\
  spect & 58.3$\pm$9.98 & 53.7$\pm$10.64 & \textbf{90.5$\pm$2.24} & 66.8$\pm$4.27 & 75.1$\pm$0.7 & 73.3$\pm$1.15 \\
  t.t.t. & 70$\pm$11.97 & 70$\pm$11.79 & \textbf{99.3$\pm$0.73} & 71.87$\pm$15.79 & 71.23$\pm$0.58 & 69.47$\pm$0.83 \\\hline

    & \multicolumn{6}{c}{$h$=3}\\\hline
  banknote & 72.85$\pm$21.49 & 72.85$\pm$5.08 & \textbf{99.09$\pm$0.26} & 64.26$\pm$19.67 & 74.76$\pm$19.7 & 91.6$\pm$0.28 \\
  blood & 76.5$\pm$0.77 & 76.68$\pm$1.36 & \textbf{79.21$\pm$1.77} & 76.21$\pm$1.03 & 77.3$\pm$1.43 & 76.21$\pm$1.03 \\
  b.c. & 0$\pm$1.46 & 0$\pm$3.9 & 0$\pm$2.64 & \textbf{89.35$\pm$8.48} & 0$\pm$0.85 & 78.5$\pm$1.51 \\
  climate & 93.68$\pm$0.9 & 81.33$\pm$27.09 & \textbf{96.15$\pm$0.59} & 95.31$\pm$4.38 & 92.44$\pm$1.36 & 93.93$\pm$0.93 \\
  fico & 50.36$\pm$2.4 & 50.36$\pm$2.4 & 56.13$\pm$4.72 & 52.18$\pm$0.11 & \textbf{71.87$\pm$0.26} & 69.7$\pm$0.28 \\
  glass & 68.25$\pm$5.65 & 68.38$\pm$4.73 & \textbf{75.25$\pm$6.9} & 37.88$\pm$2.82 & 54.38$\pm$17.82 & 61.38$\pm$6.87 \\
  image & 14.41$\pm$0.58 & 14.41$\pm$0.58 & 21.3$\pm$6.61 & 32.8$\pm$6.38 & 41.18$\pm$25.22 & \textbf{43.73$\pm$0.41} \\
  ion & 0$\pm$13.53 & 0$\pm$1.16 & 0$\pm$1.91 & \textbf{100$\pm$0} & 0$\pm$3.95 & 90.8$\pm$0.82 \\
  iris & \textbf{99.46$\pm$0.8} & 98.75$\pm$1.35 & 97.32$\pm$0.89 & 87.86$\pm$27.15 & 65.8$\pm$28.93 & 96.25$\pm$1.32 \\
  monk1 & 78.49$\pm$9.96 & 80.86$\pm$8.58 & 97.42$\pm$2.91 & \textbf{100$\pm$0} & 92.47$\pm$0.72 & 74.41$\pm$3.35 \\
  parkin & 79.73$\pm$1.15 & 81.37$\pm$5.61 & \textbf{93.15$\pm$3.59} & 89.45$\pm$9.46 & 85.21$\pm$10.3 & 88.36$\pm$0.97 \\
  soy & \textbf{100$\pm$0} & \textbf{100$\pm$0} & 90.86$\pm$15.83 & \textbf{100$\pm$0} & \textbf{100$\pm$0} & 82.29$\pm$4.24 \\
  spect & 0$\pm$4.07 & 0$\pm$1.86 & 0$\pm$3.03 & 74.4$\pm$8.58 & 0$\pm$0.92 & \textbf{75.7$\pm$2.02} \\
  t.t.t. & 55.88$\pm$15.9 & 61.11$\pm$15.79 & \textbf{98.66$\pm$0.51} & 85.43$\pm$19.47 & 78.72$\pm$0.25 & 70.53$\pm$1.45 \\\hline

    & \multicolumn{6}{c}{$h$=4}\\\hline
  banknote & 0$\pm$5.08 & 0$\pm$5.08 & 0$\pm$0.46 & 64.12$\pm$20.07 & 0$\pm$21.43 & \textbf{94.21$\pm$0.57} \\
  blood & 65.54$\pm$23.72 & 65.43$\pm$23.72 & 78.68$\pm$1.42 & 76.21$\pm$1.03 & 78$\pm$2.08 & \textbf{79.43$\pm$1.32} \\
  b.c. & 65.14$\pm$20.19 & 64.11$\pm$20.12 & \textbf{92.62$\pm$2.07} & 83.18$\pm$3.19 & 87.94$\pm$0.48 & 79.81$\pm$1.6 \\
  climate & 94.96$\pm$1.76 & 93.78$\pm$1.59 & \textbf{97.48$\pm$2.67} & 95.16$\pm$4.49 & 93.7$\pm$2.6 & 95.41$\pm$0.57 \\
  fico & 50.36$\pm$2.4 & 50.36$\pm$2.4 & 52.18$\pm$0.11 & 52.18$\pm$0.11 & \textbf{72.53$\pm$0.29} & 70.91$\pm$0.52 \\
  glass & 0$\pm$6.27 & 0$\pm$25.17 & 0$\pm$6.64 & 36.88$\pm$3.48 & 0$\pm$22.55 & \textbf{71.88$\pm$2.3} \\
  image & 14.41$\pm$0.58 & 14.41$\pm$0.58 & 24.36$\pm$11.8 & 24.02$\pm$8.16 & 48.78$\pm$33.16 & \textbf{59.83$\pm$4.96} \\
  ion & 72.62$\pm$14.29 & 66.31$\pm$13.92 & \textbf{95.44$\pm$1.34} & 78.78$\pm$19.38 & 93.46$\pm$4.35 & 92.47$\pm$0.32 \\
  iris & 98.57$\pm$1.35 & \textbf{99.29$\pm$1.6} & 97.5$\pm$0.75 & 93.04$\pm$15.57 & 69.02$\pm$32.31 & 97.14$\pm$1.32 \\
  monk1 & 84.73$\pm$6.2 & 81.29$\pm$13.72 & 95.7$\pm$5.54 & \textbf{100$\pm$0} & \textbf{100$\pm$0} & 83.44$\pm$6.65 \\
  parkin & 0$\pm$25.69 & 0$\pm$9.35 & 0$\pm$3.95 & 84.11$\pm$14.52 & 0$\pm$12.73 & \textbf{94.66$\pm$1.77} \\
  soy & \textbf{100$\pm$0} & \textbf{100$\pm$0} & 93.71$\pm$6.52 & \textbf{100$\pm$0} & \textbf{100$\pm$0} & \textbf{100$\pm$0} \\
  spect & 69.7$\pm$2.25 & 64.9$\pm$1.44 & \textbf{89$\pm$2.67} & 76$\pm$6.63 & 85$\pm$0.75 & 76.8$\pm$1.52 \\
  t.t.t. & 65.04$\pm$14.04 & 79.81$\pm$14.04 & \textbf{98.77$\pm$0.78} & 64.74$\pm$1.38 & 86.85$\pm$0.42 & 75.65$\pm$0.92 \\\hline

    & \multicolumn{6}{c}{$h$=5}\\\hline
  banknote & 55.26$\pm$5.08 & 70.55$\pm$5.08 & \textbf{99.42$\pm$0.42} & 73.16$\pm$24.51 & 77.51$\pm$22.6 & 97.03$\pm$1.06 \\
  blood & 65.43$\pm$23.72 & 65.43$\pm$23.72 & 78.07$\pm$0.88 & 76.21$\pm$1.03 & 78.14$\pm$2.21 & \textbf{80.43$\pm$1.08} \\
  b.c. & 72.43$\pm$2.52 & 71.96$\pm$1.58 & 90.28$\pm$1.38 & 93.46$\pm$6.71 & \textbf{94.11$\pm$0.67} & 82.15$\pm$1.21 \\
  climate & 0$\pm$1.07 & 0$\pm$1.3 & 0$\pm$1.26 & 92.1$\pm$0.82 & 0$\pm$3.97 & \textbf{97.19$\pm$0.79} \\
  fico & 50.36$\pm$2.4 & 50.36$\pm$2.4 & 52.18$\pm$0.11 & 52.18$\pm$0.11 & \textbf{73.3$\pm$0.25} & 71.78$\pm$0.53 \\
  glass & 31.13$\pm$2.07 & 44.5$\pm$27.94 & 73.63$\pm$5.01 & 36.13$\pm$2.09 & 64.06$\pm$27.88 & \textbf{77$\pm$1.84} \\
  image & 14.41$\pm$0.58 & 14.41$\pm$0.58 & 16.38$\pm$2.24 & 15.08$\pm$0.08 & 53.61$\pm$38.19 & \textbf{72.7$\pm$5.23} \\
  ion & 92.55$\pm$16.24 & \textbf{100$\pm$0} & 94.22$\pm$3.34 & 85.86$\pm$19.38 & 95.55$\pm$4.63 & 95.36$\pm$0.56 \\
  iris & 0$\pm$2.63 & 0$\pm$0.8 & 0$\pm$1.72 & 60.89$\pm$35.71 & 0$\pm$32.49 & \textbf{98.93$\pm$0.75} \\
  monk1 & 87.31$\pm$2.33 & 86.45$\pm$7.4 & 96.77$\pm$2.84 & \textbf{100$\pm$0} & \textbf{100$\pm$0} & 84.95$\pm$4.09 \\
  parkin & 81.37$\pm$6.63 & 84.38$\pm$7.5 & 94.66$\pm$4.06 & \textbf{100$\pm$0} & 87.74$\pm$12.94 & 97.81$\pm$2.34 \\
  soy & \textbf{100$\pm$0} & \textbf{100$\pm$0} & 94.86$\pm$7.11 & \textbf{100$\pm$0} & \textbf{100$\pm$0} & \textbf{100$\pm$0} \\
  spect & 61.4$\pm$1.56 & 66.3$\pm$7.24 & 88.4$\pm$1.08 & 82.5$\pm$11.05 & \textbf{91.8$\pm$0.98} & 80.3$\pm$1.15 \\
  t.t.t. & 0$\pm$16.3 & 0$\pm$15.14 & 0$\pm$14.86 & \textbf{100$\pm$0} & 0$\pm$0.29 & 84.09$\pm$0.89 \\\hline
\end{tabular}  
\end{table}

\begin{table}[H]
\setlength{\tabcolsep}{2.5pt}
\centering
\parbox{.48\linewidth}{
\fontsize{8}{9}\selectfont
\caption{$\mcutone$ balance vs\\imbalanced vs~$\soct$ MIS metrics.}
\begin{tabular}[H]{|lcccc:c|}\hline
    Dataset & MIS$_o$ & MIS$_b$ & Diff. Time & Diff. Acc. & $\soct$\\\hline 
  
  & \multicolumn{5}{c}{$h$=2}\\\hline
  bank & 15802.4 & 16714.2 & MEM & 0.17 & 12568.6 \\
  blood & 12098.2 & 16497.8 & 80.48 & -42.99 & 13655.4 \\
  b.c. & 16080.4 & 7689.4 & -317.18 & 3.19 & 26439.8 \\
  climate & 8505 & 17644.8 & 382.61 & -6.96 & 14286.6 \\
  fico & 0   & 900.7 & 899.75 & -2.93 & 2350.4 \\
  glass & 16762.6 & 18719.7 & -84.81 & -3.89 & 23739 \\
  image & 3928 & 4552.7 & 45.89 & 8.08 & 12311.6 \\
  ion & 3799.4 & 2745.4 & 23.12 & -11.48 & 11506.8 \\
  iris & 1428 & 2554.3 & 3.91 & 0   & 993.6 \\
  monk1 & 489.2 & 925.1 & 5.79 & 3.87 & 10120.4 \\
  parkin & 6141.4 & 6355.1 & 26.7 & -7.76 & 8864.8 \\
  soy & 0   & 0   & -0.02 & 0   & 0 \\
  spect & 8802.6 & 17973.4 & 263.48 & 0.6 & 26685.6 \\
  t.t.t. & 5076.8 & 9249.9 & 358.9 & -10.42 & 8546.2 \\\hline

    & \multicolumn{5}{c}{$h$=3}\\\hline
  bank & 6587.8 & 26586.9 & 461.89 & -16.88 & 19794.4 \\
  blood & 23204.6 & 32902.9 & (11315.96) & -12.99 & 32534.8 \\
  b.c. & 11287.6 & 16877.8 & 204.41 & -8.61 & 22547.8 \\
  climate & 10079.8 & 23466.4 & 418.95 & -39.19 & 14493.2 \\
  fico & 0   & 354.9 & MEM & -0.68 & 3167 \\
  glass & 32090 & 38070.3 & (40.53) & -2.59 & 22671.2 \\
  image & 6317.4 & 6122 & MEM & 1.11 & 6552.6 \\
  ion & 461.6 & 11556.6 & 695.84 & -7.05 & 841.8 \\
  iris & 6217.6 & 22340 & 604.46 & -6.58 & 4666.6 \\
  monk1 & 1474.2 & 951.2 & -3.12 & -4.52 & 2880.4 \\
  parkin & 1218.8 & 19181.2 & 584.75 & -2.24 & 19150.2 \\
  soy & 0   & 0   & -0.03 & -3.33 & 0 \\
  spect & 19879.6 & 24202 & (377.94) & -4.33 & 26998.4 \\
  t.t.t. & 7624.2 & 12981.7 & 176.42 & -3.79 & 11114.2 \\\hline

    & \multicolumn{5}{c}{$h$=4}\\\hline
  bank & 27087.2 & 16405 & MEM & -10.61 & 19188.4 \\
  blood & 21312 & 40788.2 & (12314.67) & -16.84 & 47543 \\
  b.c. & 13781 & 13689.6 & 79.49 & -12.5 & 23185.6 \\
  climate & 22961.6 & 27880 & 102.89 & -7.93 & 16315.2 \\
  fico & 0   & 0   & MEM & -2.93 & 3184.6 \\
  glass & 35906.8 & 50301.6 & (366.98) & -10.93 & 29072.8 \\
  image & 3522.4 & 5333.6 & MEM & 0.12 & 9320.2 \\
  ion & 12371.2 & 9233.1 & -304.6 & 1.59 & 11070.8 \\
  iris & 16674.2 & 29886.6 & 169.93 & -18.42 & 4691.4 \\
  monk1 & 870.4 & 1068.6 & 8.37 & 4.52 & 3235.2 \\
  parkin & 12356.8 & 14649.6 & 134.32 & -15.92 & 15624.2 \\
  soy & 0   & 0   & -0.07 & 0   & 0 \\
  spect & 9521.6 & 18802.1 & 352.37 & 3.28 & 22964.2 \\
  t.t.t. & 16423.4 & 7680.4 & MEM & -22 & 21234.8 \\\hline

  & \multicolumn{5}{c}{$h$=5}\\\hline
  bank & 6096 & 16776.3 & MEM & -10.61 & 14204.6 \\
  blood & 14184.6 & 22952.1 & (8982.56) & -27.65 & 52924 \\
  b.c. & 14558.2 & 10109.4 & MEM & -13.33 & 18438 \\
  climate & 14457.4 & 18057.8 & (-331.7) & -47.26 & 20489.2 \\
  fico & 0   & 0   & MEM & -2.93 & 2363.8 \\
  glass & 39421.6 & 55734 & (2190.01) & -8.52 & 38373 \\
  image & 156.4 & 4183.3 & MEM & -0.45 & 9653.2 \\
  ion & 776.4 & 10566 & 746.12 & -4.89 & 7031.8 \\
  iris & 15371 & 33595.7 & 334.17 & -37.89 & 16917.2 \\
  monk1 & 3503.4 & 833.3 & -44.68 & -8.06 & 3230.4 \\
  parkin & 18498 & 18007.2 & 150.35 & -16.33 & 8788.8 \\
  soy & 0   & 0   & -0.11 & 0.36 & 0 \\
  spect & 16869.6 & 15973.7 & (4471.14) & -10.6 & 20884 \\
  t.t.t. & 6395 & 8570 & MEM & -22 & 370.4 \\\hline
\end{tabular}\label{tab:mcut1balance}}
\parbox{.48\linewidth}{
\fontsize{8}{9}\selectfont
\caption{$\mcuttwo$ balance vs\\imbalanced vs~$\soct$ MIS metrics.}
\begin{tabular}[H]{|lcccc:c|}\hline
    Dataset & MIS$_o$ & MIS$_b$ & Diff. Time & Diff. Acc. & $\soct$ \\\hline 

  & \multicolumn{5}{c}{$h$=2}\\\hline
  bank & 15668 & 16503.2 & MEM & 0.15 & 12568.6 \\
  blood & 12781.8 & 16123.2 & 169.7 & -42.25 & 13655.4 \\
  b.c. & 14326.6 & 9873.4 & -214.85 & 1.39 & 26439.8 \\
  climate & 11139.8 & 18023.6 & 268.91 & 0.74 & 14286.6 \\
  fico & 0   & 893.9 & 899.54 & -2.93 & 2350.4 \\
  glass & 12247 & 19598.4 & 82.62 & -6.3 & 23739 \\
  image & 3773.6 & 4758.9 & 41.53 & 5.72 & 12311.6 \\
  ion & 3530.6 & 4198.7 & 61.37 & 3.64 & 11506.8 \\
  iris & 1428 & 1796.4 & -2.67 & 3.16 & 993.6 \\
  monk1 & 510.4 & 1003.1 & 6.97 & 0.97 & 10120.4 \\
  parkin & 6186.6 & 5866.2 & 21.07 & -10 & 8864.8 \\
  soy & 0   & 0   & -0.02 & 0   & 0 \\
  spect & 6067.4 & 17074.3 & 365.95 & 4.93 & 26685.6 \\
  t.t.t. & 5110 & 9167.7 & 358.86 & -11.08 & 8546.2 \\\hline

    & \multicolumn{5}{c}{$h$=3}\\\hline
  bank & 8211.2 & 25867.4 & 354.2 & -7.61 & 19794.4 \\
  blood & 20153.2 & 32333.1 & (5277.54) & -22.73 & 32534.8 \\
  b.c. & 11940 & 19711 & 309.61 & -14.31 & 22547.8 \\
  climate & 9946.8 & 21928.9 & 353.97 & -38.89 & 14493.2 \\
  fico & 0   & 355.6 & MEM & -0.68 & 3167 \\
  glass & 32755.2 & 40169.2 & (23.56) & -7.04 & 22671.2 \\
  image & 6265 & 6447.1 & MEM & -0.45 & 6552.6 \\
  ion & 605.6 & 11954.7 & 685.93 & -2.05 & 841.8 \\
  iris & 8641.8 & 25504.9 & 496.47 & -6.58 & 4666.6 \\
  monk1 & 1662.2 & 1121.1 & -1.36 & 3.55 & 2880.4 \\
  parkin & 5954.6 & 16406.7 & 323.63 & -14.69 & 19150.2 \\
  soy & 0   & 0   & -0.04 & -3.33 & 0 \\
  spect & 16591.8 & 25257.6 & 35.81 & -6.87 & 26998.4 \\
  t.t.t. & 7537.8 & 13276.3 & 176.42 & -6.92 & 11114.2 \\\hline
  
  & \multicolumn{5}{c}{$h$=4}\\\hline
  bank & 27842.6 & 15913.7 & MEM & -10.61 & 19188.4 \\
  blood & 21874 & 38713.4 & MEM & -16.84 & 47543 \\
  b.c. & 13955.4 & 13435.8 & 120.53 & -11.25 & 23185.6 \\
  climate & 26658.6 & 27490.2 & (4247.61) & -25.11 & 16315.2 \\
  fico & 0   & 0   & MEM & -2.93 & 3184.6 \\
  glass & 36539.6 & 49989 & (448.09) & 0.37 & 29072.8 \\
  image & 3081.2 & 5459.6 & MEM  & 1.47 & 9320.2 \\
  ion & 11387.6 & 9063.6 & -236.65 & 2.73 & 11070.8 \\
  iris & 15332 & 29534.8 & 289.22 & -26.05 & 4691.4 \\
  monk1 & 1288.8 & 1090.8 & 1.19 & 4.52 & 3235.2 \\
  parkin & 14276.2 & 14702.5 & 50.58 & -14.9 & 15624.2 \\
  soy & 0   & 0   & -0.04 & 0   & 0 \\
  spect & 9841.8 & 17568.2 & 408.54 & -1.94 & 22964.2 \\
  t.t.t. & 16484.4 & 7535.4 & MEM & -22 & 21234.8 \\\hline
  
  & \multicolumn{5}{c}{$h$=5}\\\hline
  bank & 28191.2 & 16325.3 & MEM & -10.61 & 14204.6 \\
  blood & 23435.8 & 22275.5 & MEM & -43.42 & 52924 \\
  b.c. & 15054.8 & 10649.9 & MEM & -11.53 & 18438 \\
  climate & 35868.4 & 17789.5 & MEM & -63.7 & 20489.2 \\
  fico & 0   & 0   & MEM & -2.93 & 2363.8 \\
  glass & 40637 & 54741.2 & (3959.71) & -6.3 & 38373 \\
  image & 0   & 4108.6 & MEM & -0.45 & 9653.2 \\
  ion & 2333.4 & 11871.8 & 700.14 & -0.23 & 7031.8 \\
  iris & 16123.4 & 26699.9 & 127.03 & -30.53 & 16917.2 \\
  monk1 & 2716 & 741 & -35.13 & -3.23 & 3230.4 \\
  parkin & 18873.2 & 18407.9 & 173.94 & -25.92 & 8788.8 \\
  soy & 0   & 0   & -0.12 & 0.83 & 0 \\
  spect & 17362.4 & 13046 & (5730.02) & -5.37 & 20884 \\
  t.t.t. & 16020.2 & 9037.2 & MEM & -22 & 370.4 \\\hline
\end{tabular}
\label{tab:mutlibalance}}
\end{table}
\normalsize

\begin{figure}[H]
\centering
    \includegraphics[width=.82\linewidth]{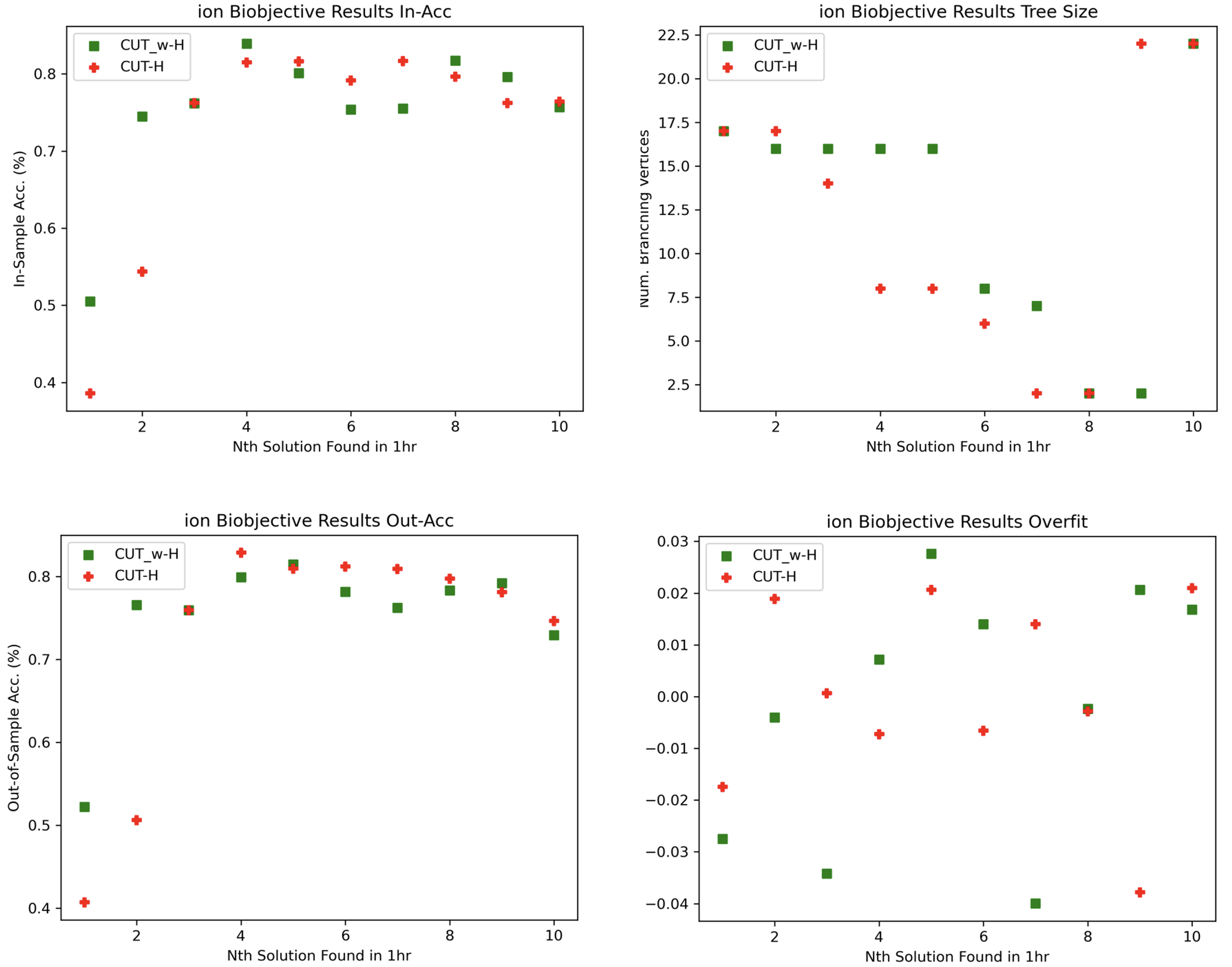}
    \vspace{-.5em}\caption{MDT biobjective results for \texttt{ion}. Priority on objective~\eqref{basemax}.}\label{fig:multi_biobj_data}
\end{figure}\vspace{-2.5em}
\begin{figure}[H]
\centering
    \includegraphics[width=.82\linewidth]{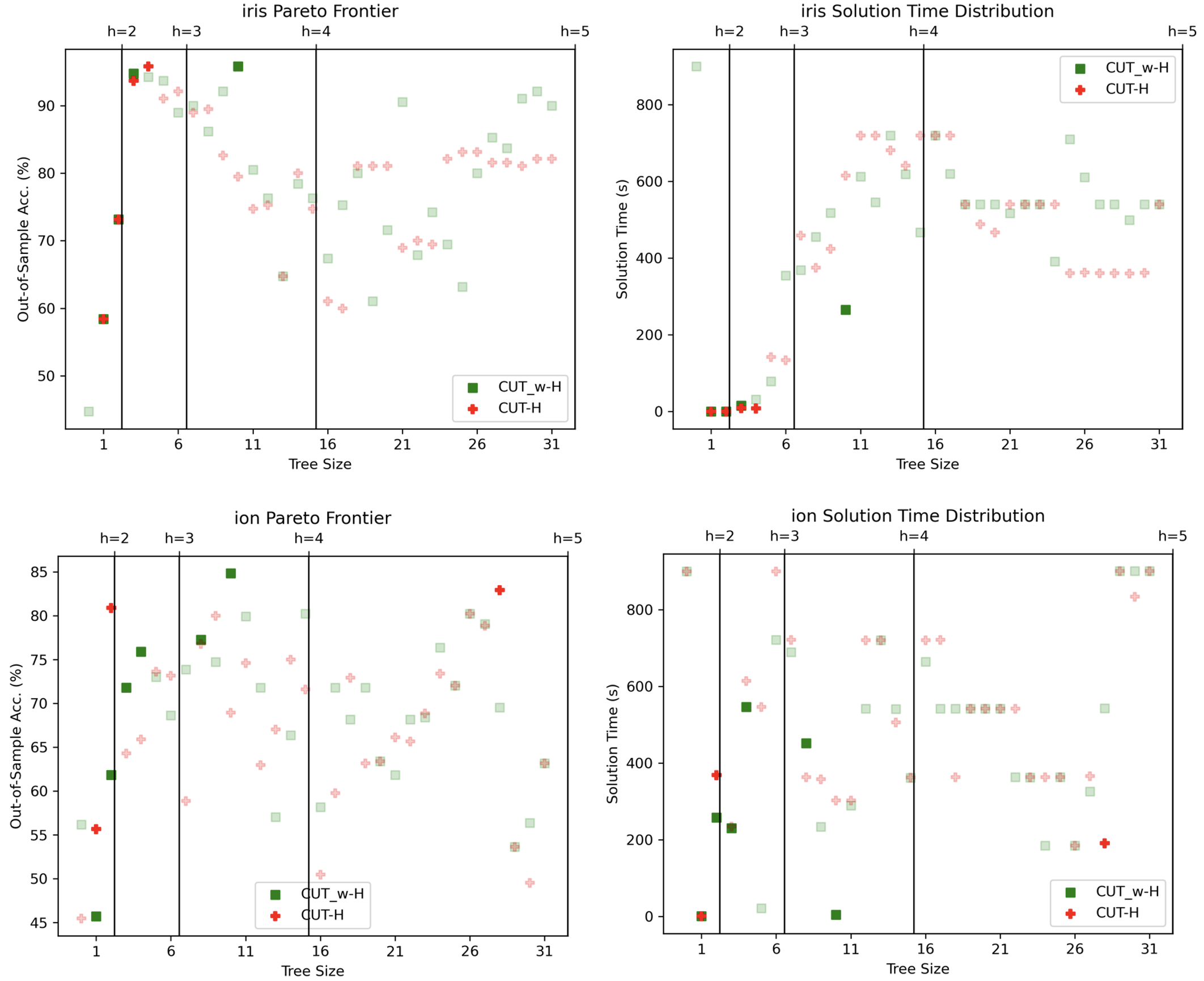}
    \vspace{-.5em}\caption{MDT Pareto Frontiers and solution time distribution for \texttt{ion} and \texttt{iris}.}\label{fig:multi_pareto}
\end{figure}

\section{Discussion}
\textbf{Optimization Comparison}
Table~\ref{tab:multi_time_gap} details the in-sample optimization performance of the models. Observe that our models report a memory crash in 11 instances, report gap in 3 instances and finds an optimal solution in 42 instances. In all the instances where our models crash the benchmark $\milo$ models all report average in-sample optimality gap values above 90$\%$. In the 42 instances where our models don't crash we either win in solution time or in-sample optimality gap in 36 such instances compared to the benchmark $\milo$ models. This suggests the strong performance of our cut-based path feasibility inequalities extends into the MDT domain. Further 7 of the 11 crashes are observed when $h=5$, the largest tree models. This is unsurprising given the known scaling issues associated with~$\milo$ decision trees and our cut-based inequalities~\eqref{cut21} increase exponentially as the tree depth increases. Of the models that do not report in-sample optimality gap many find solutions well within the 15 min TL, suggesting the models that do crash spend time introduce the shattering inequalities in an adhoc manner leading to the memory crash. As expected the branch and bound models find solutions very quickly, a few exceptions for depth $h=5$ trees in DL8.5.

\textbf{Accuracy Comparison}
Table~\ref{tab:multi_accuracy} details the out-of-sample accuracy performance of the models. Observe that our models outperform the benchmark~$\milo$ model~$\soct$ in 16 test instances by an average of 7.41$\%$, benchmark~$\octh$ in 14 instances by an average of 7.61$\%$, and branch and bound models DL8.5 and CART in 13 instances by an average of 4.17$\%$. In any average comparison $\mcuttwo$ outpeforms $\mcutone$, highlighting again the strength of our descendant based cutting vertices we find in our fractional separation procedure. The low number of instances in which our models win in out-of-sample accuracy is not a phenomena unique to only our models. Observe that $\soct$ ($\octh$) out perform the branch and bound models in only 14 (9) instances by an average of 9.80$\%$ (7.22$\%$). Similarly the margins by which the branch and bound models out perform out models (the benchmark $\milo$ models) by an average of 15.01$\%$ (13.85$\%$). Thus it is quite clear the similar performance of the $\milo$ models with respect to out-of-sample accuracy.

Table~\ref{tab:mutli_in_accuracy} details the in-sample accuracy performance of the models. Observe similar number of wins and losses between comparisons of models very close to the values of the out-of-sample comparisons. The only difference comes in the magnitude of difference between model performance. Our models lose to the benchmark $\milo$ models by an average of 15.66$\%$ and the branch and bound models by 30.74$\%$. The benchmark models lose by an average of 36.15$\%$ to the branch and bound models. The large variance in in-sample accuracy performance for our models comes from not potentially generating all the shattering inequalities~\eqref{operative} needed to properly define each hyperplane. Thus we truly observe weak separation of the data, as previously mentioned in Section~\ref{subprocesses}.

\textbf{Subprocesses Performance}
Tables~\ref{tab:mcut1balance} and~\ref{tab:mutlibalance} summarize the performance of shattering inequalities in balanced vs imbalanced trees. $MIS_o,~MIS_b$ reports the number of shattering inequalities added in imbalanced, balanced trees, respectively. Diff. Time reports the increase (decrease) in solution time of balanced over imbalanced trees; similar for Diff. Acc. Lastly, $\soct$ reports the number of $MIS$ inequalities added by $\soct$. Observe that balanced trees find more shattering inequalities in 38 (34) of the 52 test instances (5 ties) for $\mcutone$ ($\mcuttwo$). This is expected due to more branching vertices existing in the optimal solutions' tree structure. Additionally the imbalanced models must spend time solving for decision variable $p$ in the base model where as values of $p$ are fixed to zero for all vertices in vertex set $B$ of $G_h$. These additional cuts results on average in a $~170$s longer solution time. However it is quite clear that the imbalanced trees perform better in out-of-sample accuracy evident by only 11 (14) balanced trees improving accuracy results compared to their imbalanced counterparts for $\mcutone$ ($\mcuttwo$). Such results suggest balanced MDTs both take longer to generate and perform inferior to imbalanced MDTs.

Observe that $\mcutone$ ($\mcuttwo$) adds more MIS cuts in the balanced formulation in 22 (19) of the test instances, most of these instances occurring in the depths $h=\{2,3\}$. In the 11 (15) instances where our models crash $\soct$ reports more MIS cuts in 9 (12) such instances, an interesting result given our on-the-fly exponential cut-based path feasibility inequalities.

\textbf{Biobjective Performance}
Figures~\ref{fig:multi_biobj_data} and~\ref{fig:multi_pareto} summarize the biobjective performance of our models on a subset of the test instances. 

In Figure~\ref{fig:multi_biobj_data} we modify objective~\eqref{basemax} to~$\min \sum_{i\in I}(1-\sum_{v\in V\setminus \{1\}} s^i_v)$ to accommodate for the rules of hierarchical multi-objective modeling of Gurobi. Each plot provides the \emph{pool of solutions} generated by Gurobi within the 1hr TL in consecutive order. We also place a 2:1 priority on objective~\eqref{basemax} over~\eqref{basemin}. Similar to the univariate case, we observe a stairstep decrease in the tree size, however we do not observe the stairstep increase in in-sample accuracy. In both accuracy plots the best solutions have low variance with respect to accuracy metrics. We also observe a somewhat random nature of overfitting, partially due to the inconsistency in in-sample accuracy. Again we stress here a set of 10 solutions are produced within the same 1hr time limit given to the models that use a weighted objective function. Further, most of these 10 solutions perform well in out-of-sample accuracy. There is a significant increase in accuracy (in-sample and out-of-sample) at the 3$^{rd}$ solution and remains relatively high there after in the rest of solutions. This result is surprising given the process that determines the hyperplanes associated with each branching vertex is not directly solved for by our modeling process.

Figure~\ref{fig:multi_pareto} promotes the notion of larger trees performing inferior to smaller trees as all dominating points are found with trees of size $\leq 11$ branching vertices. While there do exist non-dominating points that have higher out-of-sample accuracy results gains in accuracy are marginal at greatly increased computational cost and loss of interpretability. Further, it is quite clear the warm-start solution from $k-1$ branching vertices is not helpful in finding an optimal solution for $k$ branching vertices. Observe an increase in solution time for \texttt{iris} and many with equivalent solution times for \texttt{ion}. Many of the individual solution times are $>500s$, which is over half of the time limit. Warm starts of solution $k$ perform poorly due for solution $k+1$ due to the need for additional shattering inequalities from the addition of a newly assigned branching vertex to the problem. 

 \section{Conclusions and Future Work}
In this manuscript we discussed $\milo$ formulations related to the optimal multivariate decision tree problem. We propose two novel mixed integer linear optimization formulations that can be expressed in two different ways. The first, the formulations can be thought of as extending the cut-based univariate formulations of~\citet{alston2023} into the multivariate domain through the use of shattering inequalities~\eqref{operative}. The second, the formulations can be thought of as the pruning aware, cut-based analog of~\citet{bertsimas2017} and~\citet{boutilier2022} who both use balanced, flow-based formulations; the former augments the decision tree similar to~\citet{aghaei2022} and the flow-based formulations of~\citet{alston2023} in the univariate domain.

We observe an improvement in solution time or in-sample optimality gap in 36 out of 56 test instances by our strongest, cut-based model~$\mcuttwo$. We improve solution times of by adding (fractional) path feasibility cuts at the root node of the branch and bound tree. Our approach remains competitive against existing $\milo$ models but performs relatively poorly against branch and bound models. An obvious limitation of our models is our generation of shattering inequalities. A 30 min time limit does not provide sufficient time needed for generating all inequalities. One feasible solution to this limitation is to solve the feasibility problem~\ref{operative} at each branching vertex in parallel. The pools of solutions generated by our biobjective approach are helpful in that we generate sets of well performing trees in the same 15min TL allotted to models that use a hyperparameter to control tree sparsity. However, it is not clear which tree is the best.

In the future we would like to find stronger shattering inequalities~\eqref{shatteringineq}. Currently we leverage the one-to-one correspondence between the support of~\eqref{operative} and ~\eqref{shatteringineq} to generate the inequalities. Such cuts in practice are weak, evident by analysis of dual weights of the inequalities. Symmetry in~$\milo$ formulations for decision trees must also be considered. For example, datapoints may find feasible paths through two trees with distinct branching assignments and equivalent class assignments (also yielding equivalent objective values). Some applicable symmetry breaking techniques are packing constraints~\citep{cornuejols2001}, asymmetric representative formulations (ARFs,~\citep{margot2010}) and hierarchical ordering of decision variables~\citep{jans2013}.

\bibliographystyle{informs2014} 
\bibliography{ref.bib} 

\end{document}